\documentclass[10pt,twocolumn,letterpaper]{article}

\usepackage{cvpr}              

\usepackage{graphicx}
\usepackage{amsmath}
\usepackage{amssymb}
\usepackage{booktabs}
\usepackage{comment}
\usepackage{multirow}
\usepackage{mathtools}
\usepackage{epsfig}

\usepackage{multirow}
\usepackage[table,xcdraw,dvipsnames]{xcolor}

\usepackage[breaklinks,colorlinks,citecolor = ForestGreen]{hyperref}

\usepackage[capitalize]{cleveref}
\crefname{section}{Sec.}{Secs.}
\Crefname{section}{Section}{Sections}
\Crefname{table}{Table}{Tables}
\crefname{table}{Tab.}{Tabs.}

\def\cvprPaperID{3006} 
\def\confName{CVPR}
\def\confYear{2023}

\begin{document}

\title{Rethinking Vision Transformer and Masked Autoencoder\\ in Multimodal Face Anti-Spoofing}

\author{Zitong Yu\textsuperscript{1}, Rizhao Cai\textsuperscript{2}, Yawen Cui\textsuperscript{3}, Xin Liu\textsuperscript{4}, Yongjian Hu\textsuperscript{5}, Alex Kot\textsuperscript{2}\\
\textsuperscript{1}Great Bay University  \qquad  \textsuperscript{2}Nanyang Technological University \qquad 
\textsuperscript{3}University of Oulu \\
\textsuperscript{4}Lappeenranta-Lahti University of Technology \qquad 
\textsuperscript{5}South China University of Technology}


\maketitle

\begin{abstract}

Recently, vision transformer (ViT) based multimodal learning methods have been proposed to improve the robustness of face anti-spoofing (FAS) systems. However, there are still no works to explore the fundamental natures (\textit{e.g.}, modality-aware inputs, suitable multimodal pre-training, and efficient finetuning) in vanilla ViT for multimodal FAS. In this paper, we investigate three key factors (i.e., inputs, pre-training, and finetuning) in ViT for multimodal FAS with RGB, Infrared (IR), and Depth. First, in terms of the ViT inputs, we find that leveraging local feature descriptors benefits the ViT on IR modality but not RGB or Depth modalities. Second, in observation of the inefficiency on direct finetuning the whole or partial ViT, we design an adaptive multimodal adapter (AMA), which can efficiently aggregate local multimodal features while freezing majority of ViT parameters. Finally, in consideration of the task (FAS vs. generic object classification) and modality (multimodal vs. unimodal) gaps, ImageNet pre-trained models might be sub-optimal for the multimodal FAS task. To bridge these gaps, we propose the modality-asymmetric masked autoencoder (M$^{2}$A$^{2}$E) for multimodal FAS self-supervised pre-training without costly annotated labels. Compared with the previous modality-symmetric autoencoder, the proposed M$^{2}$A$^{2}$E is able to learn more intrinsic task-aware representation and compatible with modality-agnostic (e.g., unimodal, bimodal, and trimodal) downstream settings. Extensive experiments with both unimodal (RGB, Depth, IR) and multimodal (RGB+Depth, RGB+IR, Depth+IR, RGB+Depth+IR) settings conducted on multimodal FAS benchmarks demonstrate the superior performance of the proposed methods. We hope these findings and solutions can facilitate the future research for ViT-based multimodal FAS.

\end{abstract}

\vspace{-1.5em}
\section{Introduction}
\vspace{-0.3em}

\thispagestyle{empty}

Face recognition technology has widely used in many intelligent systems due to their convenience and remarkable accuracy. However, face recognition systems are still vulnerable to presentation attacks (PAs) ranging from print, replay and 3D-mask attacks. Therefore, both the academia and industry have recognized the critical role of face anti-spoofing (FAS) for securing the face recognition system.


In the past decade, plenty of hand-crafted features based~\cite{boulkenafet2015face,Boulkenafet2017Face,Komulainen2014Context,Patel2016Secure} and deep learning based~\cite{qin2019learning,Liu2018Learning,yang2019face,Atoum2018Face,Gan20173D,george2019deep} methods have been proposed for unimodal FAS. Despite satisfactory performance in seen attacks and environments, unimodal methods generalize poorly on emerging novel attacks and unseen deployment conditions. Thanks to the advanced sensors with various modalities (e.g., RGB, Infrared (IR), Depth, Thermal)~\cite{george2019biometric}, multimodal methods facilitate the FAS applications under some high-security scenarios with low false acceptance errors (e.g., face payment and vault entrance guard).

Recently, due to the strong long-range and cross-modal representation capacity, vision transformer (ViT)~\cite{dosovitskiy2020image} based methods~\cite{liuma,george2020effectiveness} have been proposed to improve the robustness of FAS systems. However, these methods focus on direct finetuning ViTs~\cite{george2020effectiveness} or modifying ViTs with complex and powerful modules~\cite{liuma}, which cannot provide enough insights on bridging the fundamental natures (e.g., modality-aware inputs, suitable multimodal pre-training, and efficient finetuning) of ViT in multimodal FAS. Despite mature exploration and finds~\cite{he2022masked,bachmann2022multimae,xiao2021early} of ViT on other computer vision communities (e.g., generic object classification~\cite{chen2022conv}), these knowledge might not be fully aligned for the multimodal FAS due to the task and modality gaps.

Compared with CNN, ViT usually aggregates the coarse intra-patch info at the very early stage and then propagates the inter-patch global attentional features. On other words, it neglects the local detailed clues for each modality. According to the prior evidence from MM-CDCN~\cite{yu2020multi}, local fine-grained features from multiples levels benefits the live/spoof clue representation in convolutional neural networks (CNN) from different modalities. \textit{Whether local descriptors/features can improve the ViT-based multimodal FAS systems is worth exploring.}

Compared with CNNs, ViTs usually have huger parameters to train, which easily overfit on the FAS task with limited data amount and diversity. Existing works show that direct finetuning the last classification head~\cite{george2020effectiveness} or training extra lightweight adapters~\cite{huang2022adaptive} can achieve better performance than fully finetuning. However, all these observations are based on the unimodal RGB inputs, it is unclear how different ViT-based transfer learning techniques perform on 1) other unimodal scenario (IR or Depth modality); and 2) multimodal scenario (e.g., RGB+IR+Depth). Moreover, \textit{to design more efficient transfer learning modules for ViT-based multimodal FAS should be considered.}

Existing multimodal FAS works usually finetune the ImageNet pre-trained models, which might be sub-optimal due to the huge task (FAS vs. generic object classification) and modality (multimodal vs. unimodal) gaps. Meanwhile, in consideration of costly collection of large-scale annotated live/spoof data, self-supervised pre-training without labels~\cite{muhammad2022self} is potential for model initialization in multimodal FAS. Although a few self-supervised pre-training methods (e.g., masked image modeling (MIM)~\cite{bachmann2022multimae,chen2022multi} and contrastive learning~\cite{akbari2021vatt}) are developed for multimodal (e.g., vision-language) applications, there are still no self-supervised pre-trained models specially for multimodal FAS. \textit{To investigate the discrimination and generalization capacity of pre-trained models and design advanced self-supervision strategies are crucial for ViT-based multimodal FAS.}

Motivated by the discussions above, in this paper we rethink the ViT-based multimodal FAS into three aspects, i.e., modality-aware inputs, suitable multimodal pre-training, and efficient finetuning. Besides the elaborate investigations, we also provide corresponding elegant solutions to 1) establish powerful inputs with local descriptors~\cite{bhattacharjee2019pattern,dalal2005histograms} for IR modality; 2) efficiently finetune multimodal ViTs via adaptive multimodal adapters; and 3) pre-train generalized multimodal model via modality-asymmetric masked autoencoder. Our contributions include:


\begin{itemize}
\setlength\itemsep{-0.1em}
\vspace{-0.5em}
    \item We are the first to investigate three key factors (i.e., inputs, pretraining, and finetuning) for ViT-based multimodal FAS. We find that 1) leveraging local feature descriptors benefits the ViT on IR modality; 2) partially finetuning or using adapters can achieve reasonable performance for ViT-based multimodal FAS but still far from satisfaction; and 3) mask autoencoder~\cite{he2022masked,bachmann2022multimae} pre-training cannot provide better finetuning performance compared with ImageNet pre-trained models.

    \item  We design the adaptive multimodal adapter (AMA) for ViT-based multimodal FAS, which can efficiently aggregate local multimodal features while freezing majority of ViT parameters.

    \item We propose the modality-asymmetric masked autoencoder (M$^{2}$A$^{2}$E) for multimodal FAS self-supervised pre-training. Compared with modality-symmetric autoencoders~\cite{he2022masked,bachmann2022multimae}, the proposed M$^{2}$A$^{2}$E is able to learn more intrinsic task-aware representation and compatible with modality-agnostic downstream settings. To our best knowledge, this is the first attempt to design the MIM framework for generalized multimodal FAS.

    \item Our proposed methods achieve state-of-the-art performance with most of the modality settings on both intra- as well as cross-dataset testings. 
    
    
\end{itemize}

\section{Related Work}


\noindent\textbf{Multimodal face anti-spoofing.}\quad      
With multimodal inputs (e.g., RGB, IR, Depth, and Thermal), there are a few multimodal FAS works that consider input-level~\cite{nikisins2019domain,liu2021data,george2020learning,nikisins2019domain} and decision-level~\cite{zhang2019feathernets} fusions. Besides, mainstream FAS methods extract complementary multi-modal features using feature-level fusion~\cite{yu2020multi,zhang2019dataset,liu2021face,wang2022conv,liuma,li2021asymmetric} strategies. As there are redundancy across multi-modal features, direct feature concatenation~\cite{yu2020multi} easily results in high-dimensional features and overiftting. To alleviate this issue, Zhang et al.~\cite{zhang2019dataset,zhang2020casia} propose a feature re-weighting mechanism to select the informative and discard the redundant channel features among RGB, IR, and Depth modalities. Shen et al.~\cite{shen2019facebagnet} design a Modal Feature Erasing operation to randomly dropout partial-modal features to prevent modality-aware overftting. George and Marcel~\cite{george2021cross} present a cross-modal focal loss to modulate the loss contribution of each modality, which benefts the model to learn complementary information among modalities.

\begin{figure*}
\centering
\vspace{-0.8em}
\includegraphics[scale=0.37]{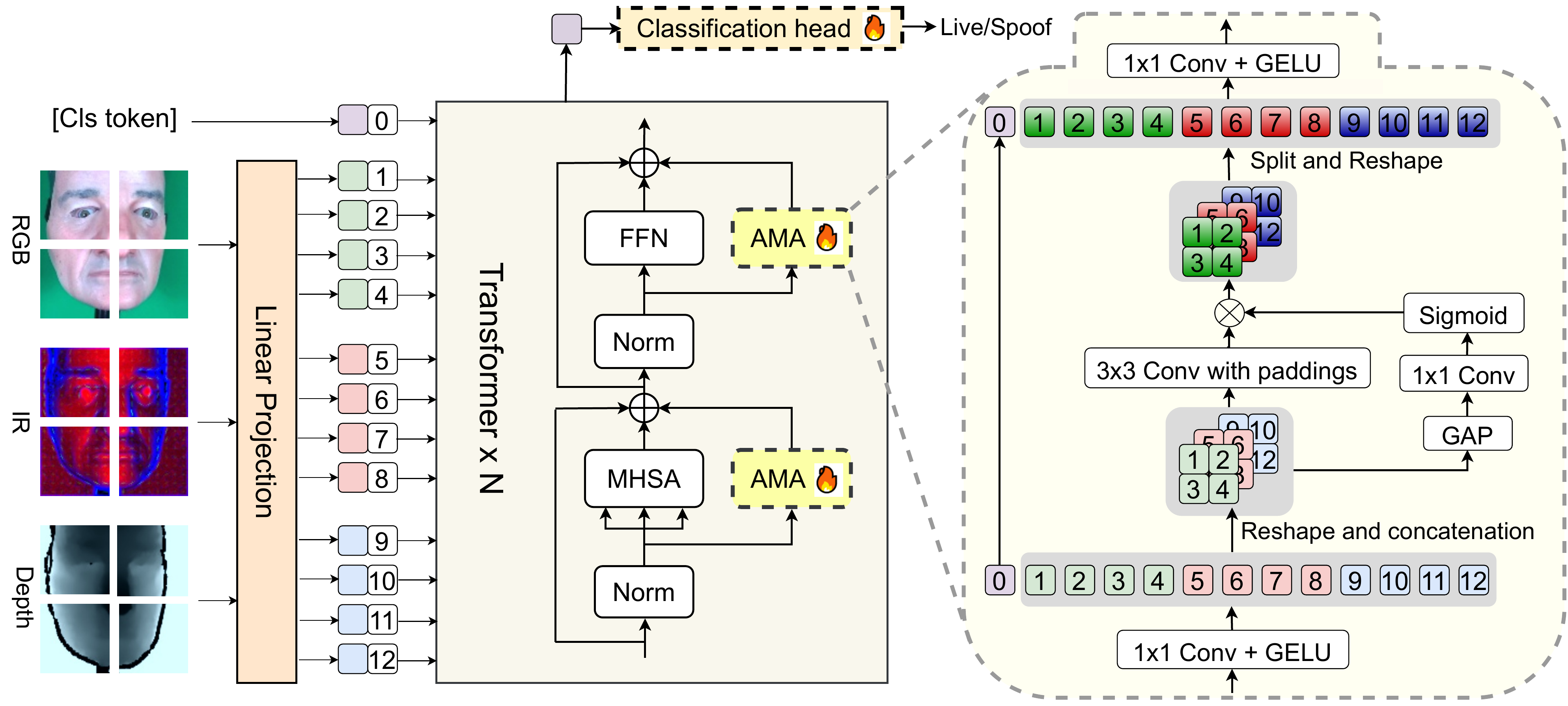}
\vspace{-1.0em}
  \caption{\small{
  Framework of the ViT finetuning with adaptive multimodal adapters (AMA). The AMA and classification head are \textbf{trainable} while the linear projection and vanilla transformer blocks are fixed with the pre-trained parameters. `MHSA', 'FFN', and 'GAP' are short for the multihead self-attention, feed-forword network, and global average pooling, respectively.
  }
}
\label{fig:AMA}
\vspace{-1.5em}
\end{figure*}

\begin{figure}
\centering
\includegraphics[scale=0.41]{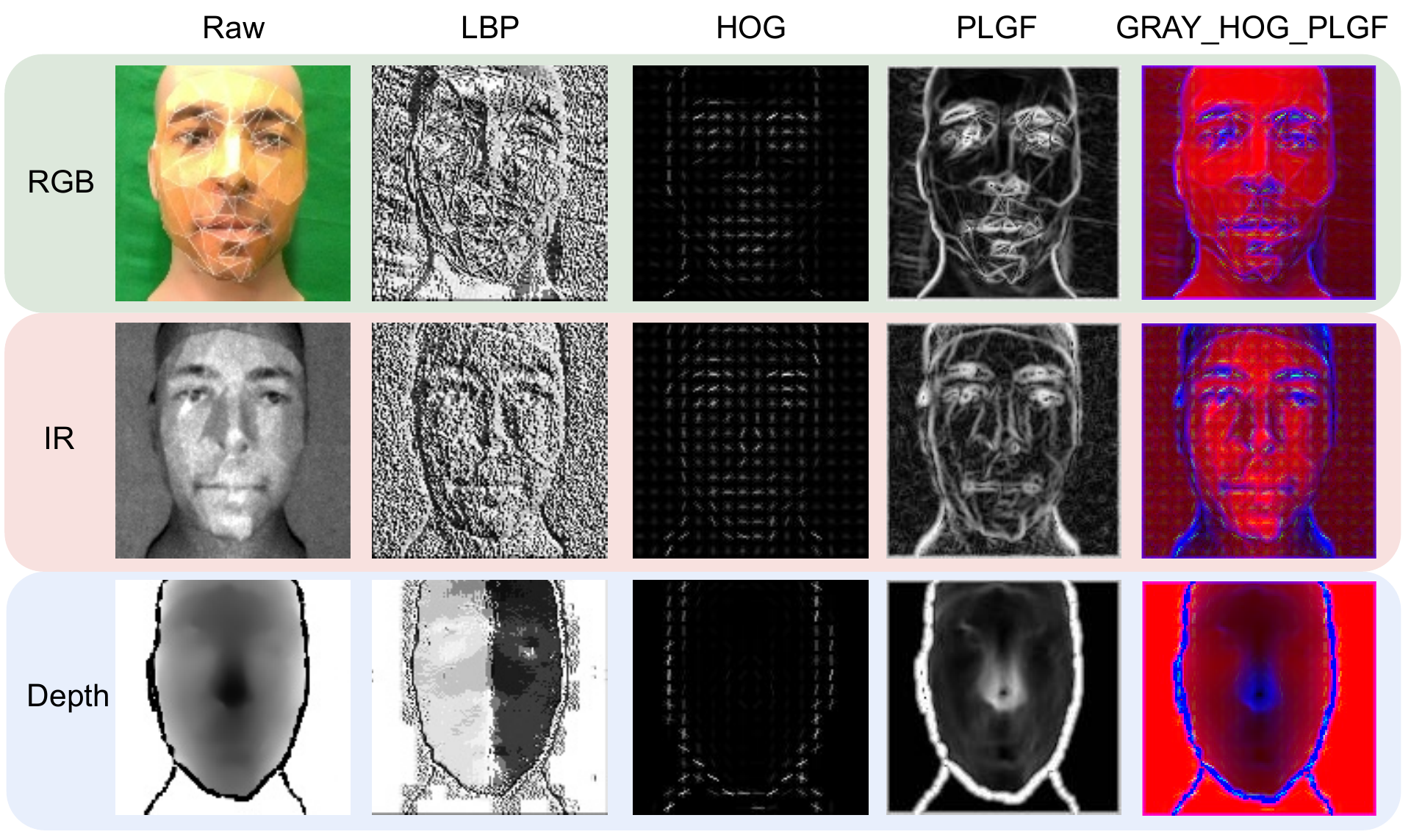}
\vspace{-1.1em}
  \caption{\small{
 Visualization of three classical local descriptors (i.e., LBP~\cite{ojala2002multiresolution}, HOG~\cite{dalal2005histograms}, and PLGF~\cite{bhattacharjee2019pattern}) and their compositions. }
  }
\label{fig:LBPHOG}
\vspace{-1.4em}
\end{figure}

\vspace{0.5em}
\noindent\textbf{Transformer for vision tasks.}\quad  
Transformer is proposed in~\cite{vaswani2017attention} to model sequential data in the field of NLP. Then ViT~\cite{dosovitskiy2020image} is proposed recently by feeding transformer with sequences of image patches for image classification. In consideration of the data hungry characteristic of ViT, direct training ViTs from scratch would result in severe overfitting. On the one hand, fast transferring (e.g., adapter~\cite{houlsby2019parameter,chen2022conv,jie2022convolutional} and prompt~\cite{zhou2022learning} tuning) while fixed most of pre-trained models' parameters is usually efficient for downstream tasks. On the other hand, self-supervised masked image modeling (MIM) methods (e.g., BEiT~\cite{bao2021beit} and MAE~\cite{he2022masked,bachmann2022multimae}) benefit the excellent representation learning, which improve the finetuning performance in downstream tasks.

Meanwhile, a few works introduce vision transformer for FAS~\cite{liuma,ming2022vitranspad,george2020effectiveness,wang2022face,wang2022learning,yu2021transrppg}. On the one hand, ViT is adopted in the spatial domain~\cite{ming2022vitranspad,george2020effectiveness,wang2022face} to explore live/spoof relations among local patches. On the other hand, global features temporal abnormity~\cite{wang2022learning} or physiological periodicity~\cite{yu2021transrppg} are extracted applying ViT in the temporal domain. Recently, Liu and Liang~\cite{liuma} develop the modality-agnostic transformer blocks to supplement liveness features for multimodal FAS. Despite convincing performance via modified ViT with complex customized modal-disentangled and cross-modal attention modules~\cite{liuma}, there are still no works to explore the fundamental natures (e.g., modality-aware inputs, suitable multimodal pre-training, and efficient finetuning) in vanilla ViT for multimodal FAS.

\section{Methodology}
\label{sec:method}
To benefit the exploration of fundamental natures of ViT for multimodal FAS, here we adopt the \textit{simple}, \textit{elegant}, and \textit{unified} ViT framework as baselines. As illustrated in the left part (without `AMA') of Fig.~\ref{fig:AMA}, the vanilla ViT consists of a patch tokenizer $\mathbf{E}_{\text{patch}}$ via linear projection, $N$ transformer blocks $\mathbf{E}^{i}_{\text{trans}}$ ($ i=1,...,N$) and a classification head $\mathbf{E}_{\text{head}}$. The unimodal ($X_{\text{RGB}}$, $X_{\text{IR}}$, $X_{\text{Depth}}$) or multimodal ($X_{\text{RGB+IR}}$, $X_{\text{RGB+Depth}}$, $X_{\text{IR+Depth}}$, $X_{\text{RGB+IR+Depth}}$) inputs are passed over $\mathbf{E}_{\text{patch}}$ to generate the visual tokens $T^\text{Vis}$, which is concatenated with learnable class token $T^\text{Cls}$, and added with position embeddings. Then all patch tokens $T^\text{All}=[T^\text{Vis},T^\text{Cls}]$ will be forwarded with $\mathbf{E}_{\text{trans}}$. Finally, $T^{\text{Cls}}$ is sent to $\mathbf{E}_{\text{head}}$ for binary live/spoof classification.

We will first briefly introduce different local descriptor based inputs in Sec.~\ref{sec:LBP}, then introduce the efficient ViT finetuning with AMA in Sec.~\ref{sec:AMA}, and at last present the generalized multimodal pre-training via  M$^{2}$A$^{2}$E in Sec.~\ref{sec:MAE}. 

\subsection{Local Descriptors for Multimodal ViT}
\label{sec:LBP}

Besides the raw multimodal inputs, we consider three local features and their compositions for multimodal ViT. The motivations behind are that the vanilla ViT with raw inputs is able to model rich cross-patch semantic contexts but sensitive to illumination and neglecting the local fine-grained spoof clues. Explicitly leveraging local descriptors as inputs might benefit multimodal ViT mining more discriminative fine-grained spoof clues~\cite{yu2020multi,yu2020searching,yu2021dual,yu2020face} as well as illumination-robust live/spoof features~\cite{li2021asymmetric}.

\vspace{0.3em}
\noindent\textbf{Local binary pattern (LBP).}\quad  
LBP~\cite{ojala2002multiresolution} computes a binary pattern via thresholding central difference among neighborhood pixels. Fine-grained textures and illumination invariance make LBP robust for generalized FAS~\cite{li2019face}. For a center pixel $I_{c}$ and a neighboring pixel $I_{i}(i=1,2,...,p)$, LBP can be formalized as follows: 
\vspace{-0.9em}
\begin{equation}
\begin{split}
\text{LBP}&=\sum_{i=1}^{p}F(I_{i}-I_{c})\times 2^{i-1},\\
F(I)&=\left\{\begin{matrix}
1, \quad\quad I\geq 0,\\ 0,\quad \text{otherwise.}
\end{matrix}\right. 
\vspace{-0.1em}
\end{split}
\end{equation}
Typical LBP maps are shown in second column of Fig.~\ref{fig:LBPHOG}.

\vspace{0.3em}
\noindent\textbf{Histograms of oriented gradients (HOG).}\quad  
HOG~\cite{dalal2005histograms} describes the distribution of gradient orientations or edge directions within a local subregion. It is implemented via firstly computing magnitudes and orientations of gradients at each pixel, and then the gradients within each small local subregion are accumulated into orientation histogram vectors of several bins, voted by gradient magnitudes. Due to the partial invariance to geometric and photometric changes, HOG features might be robust for the illumination-sensitive modalities like RGB and IR. The visualization results are shown in third column of Fig.~\ref{fig:LBPHOG}.

\vspace{0.3em}
\noindent\textbf{Pattern of local gravitational force (PLGF).}\quad  
Inspired by Law of Universal Gravitation, PLGF~\cite{bhattacharjee2019pattern} describes the image interest regions via local gravitational force magnitude, which is useful to reduce the impact of illumination/noise variation while preserving edge-based low-level clues. It can be formulated as:
\vspace{-0.5em}
\begin{equation}
\begin{split}
\text{PLGF}&=arctan(\sqrt{(\frac{I*M_{x}}{I})^{2}+(\frac{I*M_{y}}{I})^{2}}),\\
M_{x}(m,n) &=\left\{\begin{matrix}
\frac{cos(arctan(m/n))}{m^{2}+n^{2}}, \quad (m^{2}+n^{2})>0,\\ 0,\quad\quad\quad\quad \text{otherwise,}
\end{matrix}\right.  \\
M_{y}(m,n) &=\left\{\begin{matrix}
\frac{sin(arctan(m/n))}{m^{2}+n^{2}}, \quad (m^{2}+n^{2})>0,\\ 0,\quad\quad\quad\quad \text{otherwise,}
\end{matrix}\right. 
\vspace{-0.1em}
\end{split}
\end{equation}
where $I$ is the raw image. $M_{x}$ and $M_{y}$ are two filter masks for gravitational force calculation. $m$ and $n$ are indexes denoting the relative position to the center. $*$ is convolution operation sliding along all pixels. The visualization of PLGF maps are shown in fourth column of Fig.~\ref{fig:LBPHOG}.

\vspace{0.3em}
\noindent\textbf{Composition.}\quad  
In consideration of the complementary characteristics from raw image and local descriptors, we also study the compositions among these features via input-level concatenation. For example, `GRAY\_HOG\_PLFG' denotes three-channel inputs (raw gray-scale channel + HOG + PLFG), which is visualized in last column of Fig.~\ref{fig:LBPHOG}.

\subsection{Adaptive Multimodal Adapter}
\label{sec:AMA}

Recent studies have verified that introducing adapters~\cite{huang2022adaptive} with fully connected (FC) layers can improve the FAS performance when training data is not adequate. However, FC-based adapter focuses on the intra-token feature refinement but neglects 1) contextual features from local neighbor tokens; and 2) multimodal features from cross-modal tokens. To tackle these issues, we extend the convolutional adapter (ConvAdapter)~\cite{jie2022convolutional} into a multimodal version for multimodal FAS.

As illustrated in Fig.~\ref{fig:AMA}, instead of directly finetuning the transformer blocks $\mathbf{E}_{\text{trans}}$, we fix all the pre-trained parameters from $\mathbf{E}_{\text{patch}}$ and $\mathbf{E}_{\text{trans}}$ while training only adaptive multimodal adapters (AMA) and $\mathbf{E}_{\text{head}}$. An AMA module consists of four parts: 1) an 1×1 convolution with GELU $\Theta_{\downarrow}$ for dimension reduction from the original channels $D$ to a hidden dimension $D'$; 2) a 3×3 2D convolution $\Theta_{\text{2D}}$ mapping channels $D'$×$K$ to $D'$ for multimodal local feature aggregation, where $K$ means the modality numbers; 3) an adaptive modality weight ($w_{1},...,w_{K}$) generator via cascading global averaging pooling (GAP), 1×1 convolution $\Theta_{\text{Ada}}$ to project channels from $D'$×$K$ to $K$, and the Sigmoid function $\sigma$; and 4) an 1×1 convolution with GELU $\Theta_{\uparrow}$ for dimension expansion to $D$. As features from different modalities are already spatially aligned, we restore the 2D structure for each modality after the channel squeezing. Similarly, the 2D structure will be flatten into 1D tokens before the channel expanding. The AMA can be formulated as
\vspace{-0.5em}
\begin{equation}
\begin{split}
&T^\text{Vis}_{\text{Kmodal}}=\text{Concat}[\Theta_{\downarrow}(T^\text{Vis}_{\text{RGB}}),\Theta_{\downarrow}(T^\text{Vis}_{\text{IR}}),\Theta_{\downarrow}(T^\text{Vis}_{\text{Depth}})], \\
&w_{\text{RGB}},w_{\text{IR}},w_{\text{Depth}} = \sigma (\Theta_{\text{Ada}}(\text{GAP}(T^\text{Vis}_{\text{Kmodal}}))), \\
&T^\text{Vis}_{\text{Kmodal}}=\Theta_{\text{2D}}(T^\text{Vis}_{\text{Kmodal}}), \\
&T^\text{Vis}_{\text{Kmodal}}=\text{Concat}[w_{\text{RGB}} \cdot T^\text{Vis}_{\text{Kmodal}}
,w_{\text{IR}} \cdot T^\text{Vis}_{\text{Kmodal}},w_{\text{Depth}} \cdot T^\text{Vis}_{\text{Kmodal}}], \\
&\text{AMA}=\text{Concat}[\Theta_{\uparrow}(\Theta_{\downarrow}(T^\text{Cls})),\Theta_{\uparrow}(T^\text{Vis}_{\text{Kmodal}})].
\vspace{-0.8em}
\end{split}
\label{eq:AMA}
\end{equation}
Here we show an example when $K$=3 (i.e., RGB+ IR+Depth) in Eq.(\ref{eq:AMA}), and AMA is flexible for arbitrary modalities (e.g., RGB+IR). Note that AMA is equivalent to vanilla ConvAdapter~\cite{jie2022convolutional} in unimodal setting when $K$=1.


\begin{figure}[t]
\centering
\includegraphics[scale=0.48]{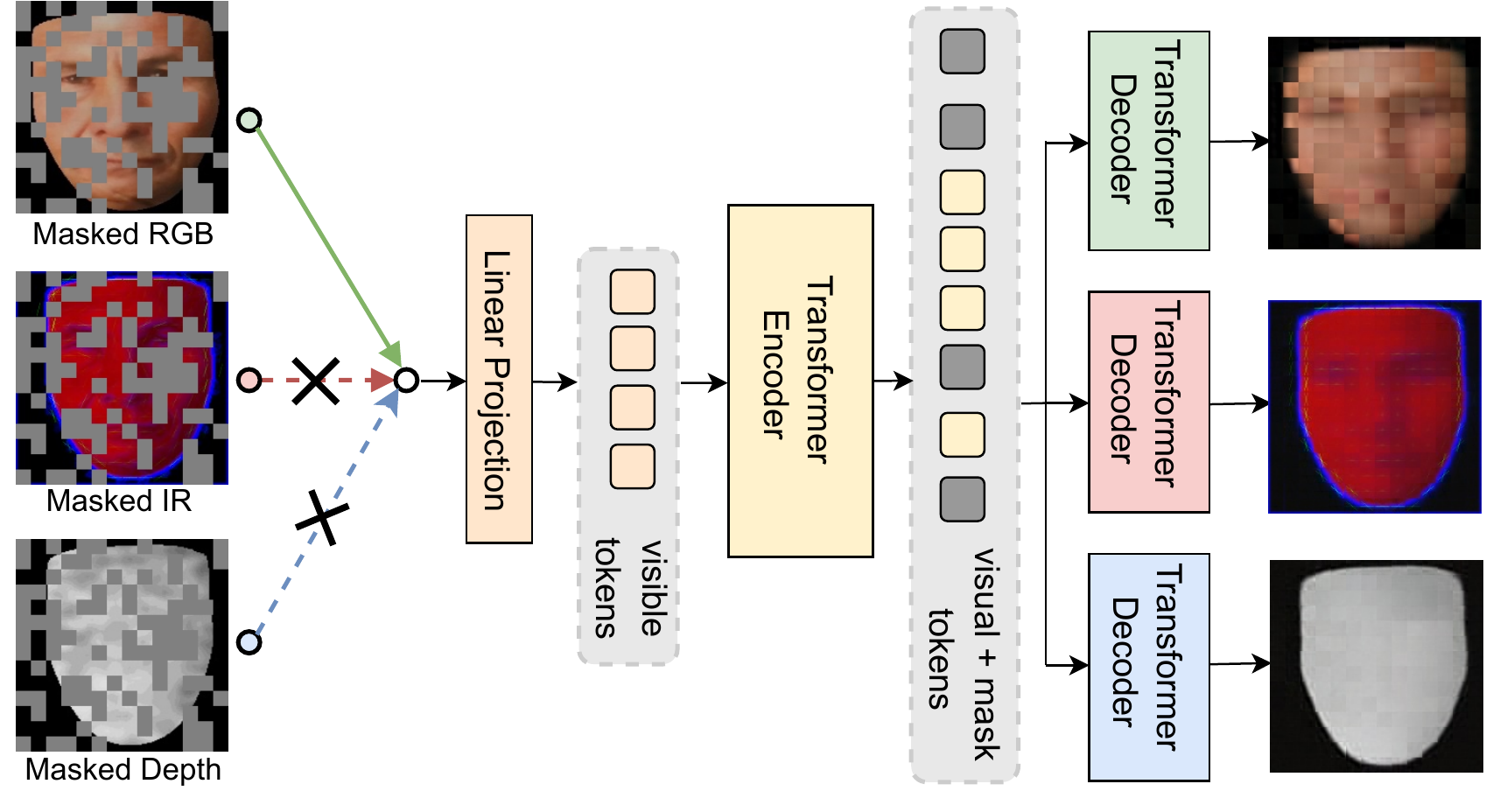}
\vspace{-2.0em}
  \caption{\small{
 The framework of the modality-asymmetric masked autoencoder (M$^{2}$A$^{2}$E). Different from previous multimodal MAE~\cite{bachmann2022multimae} masking all modalities as inputs, our M$^{2}$A$^{2}$E randomly selects unimodal masked input for multimodal reconstruction.}
  }
\label{fig:MAE}
\vspace{-0.8em}
\end{figure}

\subsection{Modality-Asymmetric Masked Autoencoder}
\label{sec:MAE}

Existing multimodal FAS works usually finetune the ImageNet pre-trained models, which might be sub-optimal due to the huge task and modality gaps. Meanwhile, in consideration of costly collection of large-scale annotated live/spoof data, self-supervised pre-training without labels~\cite{muhammad2022self} is potential for model initialization in multimodal FAS. Here we propose the modality-asymmetric masked autoencoder (M$^{2}$A$^{2}$E) for multimodal FAS self-supervised pre-training.

\begin{table*}[t]\small
\centering
\caption{ACER(\%) results of protocols `seen' and `unseen' on WMCA. The values ACER(\%) reported on testing sets are obtained with thresholds computed for BPCER=1\% on development sets. Best results are marked in \textbf{bold}. `ConvA' indicates the ConvAdapter~\cite{jie2022convolutional}. } \label{tab:WMCA}

\vspace{-0.9em}

\resizebox{0.89\textwidth}{!}{\begin{tabular}{|lccccccccc|}
\hline
\multicolumn{1}{|c|}{}                                  & \multicolumn{1}{c|}{}                                     & \multicolumn{8}{c|}{\textbf{Unseen}}                                                                                                                   \\ \cline{3-10} 
\multicolumn{1}{|c|}{\multirow{-2}{*}{\textbf{Method}}} & \multicolumn{1}{c|}{\multirow{-2}{*}{\textbf{Seen}}}      & \multicolumn{1}{c|}{\textbf{Flexiblemask}}        & \multicolumn{1}{c|}{\textbf{Replay}}             & \multicolumn{1}{c|}{\textbf{Fakehead}}           & \multicolumn{1}{c|}{\textbf{Prints}}                      & \multicolumn{1}{c|}{\textbf{Glasses}}             & \multicolumn{1}{c|}{\textbf{Papermask}}          & \multicolumn{1}{c|}{\textbf{Rigidmask}}                   & \textbf{mean±std}                           \\ \hline
\multicolumn{10}{|c|}{\cellcolor[HTML]{EFEFEF}Modality: \textbf{RGB}}                                                                                                                   \\ \hline
\multicolumn{1}{|l|}{MC-CNN~\cite{george2019biometric}}                            & \multicolumn{1}{c|}{32.82}                                & \multicolumn{1}{c|}{22.80}                        & \multicolumn{1}{c|}{31.40}                       & \multicolumn{1}{c|}{\textbf{1.90}}               & \multicolumn{1}{c|}{30.00}                                & \multicolumn{1}{c|}{50.00}                        & \multicolumn{1}{c|}{\textbf{4.80}}               & \multicolumn{1}{c|}{18.30}                                & 22.74±15.33                                 \\ \hline
\multicolumn{1}{|l|}{CCL(ResNet50)~\cite{liu2021contrastive}}                     & \multicolumn{1}{c|}{30.69}                                & \multicolumn{1}{c|}{\textbf{4.76}}                & \multicolumn{1}{c|}{15.37}                       & \multicolumn{1}{c|}{24.67}                       & \multicolumn{1}{c|}{19.03}                                & \multicolumn{1}{c|}{16.80}                        & \multicolumn{1}{c|}{9.51}                        & \multicolumn{1}{c|}{17.62}                                & 15.39±6.51                                           \\ \hline
\multicolumn{1}{|l|}{CCL(CDCN)~\cite{liu2021contrastive}}                         & \multicolumn{1}{c|}{27.14}                                & \multicolumn{1}{c|}{7.18}                         & \multicolumn{1}{c|}{11.79}                       & \multicolumn{1}{c|}{21.82}                       & \multicolumn{1}{c|}{20.53}                                & \multicolumn{1}{c|}{35.13}                        & \multicolumn{1}{c|}{18.91}                       & \multicolumn{1}{c|}{15.10}                                & 18.64±8.91                                  \\ \hline
\multicolumn{1}{|l|}{ViT~\cite{dosovitskiy2020image}}                               & \multicolumn{1}{c|}{{\color[HTML]{000000} 9.84}}          & \multicolumn{1}{c|}{18.4}                         & \multicolumn{1}{c|}{19.94}                       & \multicolumn{1}{c|}{13.67}                       & \multicolumn{1}{c|}{1.92}                                 & \multicolumn{1}{c|}{24.58}                        & \multicolumn{1}{c|}{5.59}                        & \multicolumn{1}{c|}{9.59}                                 & 13.38+-8.17                                 \\ \hline
\multicolumn{1}{|l|}{\textbf{ViT+ConvA (Ours)}}         & \multicolumn{1}{c|}{{\color[HTML]{000000} \textbf{4.72}}} & \multicolumn{1}{c|}{18.64}                        & \multicolumn{1}{c|}{\textbf{10.07}}              & \multicolumn{1}{c|}{7.98}                        & \multicolumn{1}{c|}{0.43}                                 & \multicolumn{1}{c|}{20.14}                        & \multicolumn{1}{c|}{6.38}                        & \multicolumn{1}{c|}{\textbf{2.32}}                        & \textbf{9.42+-7.56}                         \\ \hline
\multicolumn{1}{|l|}{\textbf{ViT+ConvA+M$^{2}$A$^{2}$E (Ours)}}     & \multicolumn{1}{c|}{8.95}                                 & \multicolumn{1}{c|}{23.68}                        & \multicolumn{1}{c|}{11.42}                       & \multicolumn{1}{c|}{2.28}                        & \multicolumn{1}{c|}{\textbf{0.29}}                        & \multicolumn{1}{c|}{27.98}                        & \multicolumn{1}{c|}{7.20}                        & \multicolumn{1}{c|}{7.65}                                 & 11.5+-10.52                                 \\ \hline
\multicolumn{10}{|c|}{\cellcolor[HTML]{EFEFEF}Modality: \textbf{IR}}                                                                                           \\ \hline
\multicolumn{1}{|l|}{RDWT-Haralick~\cite{george2019biometric}}                     & \multicolumn{1}{c|}{6.26}                                 & \multicolumn{1}{c|}{-}                            & \multicolumn{1}{c|}{-}                           & \multicolumn{1}{c|}{-}                           & \multicolumn{1}{c|}{-}                                    & \multicolumn{1}{c|}{-}                            & \multicolumn{1}{c|}{-}                           & \multicolumn{1}{c|}{-}                                    & -                                           \\ \hline
\multicolumn{1}{|l|}{MC-CNN~\cite{george2019biometric}}                            & \multicolumn{1}{c|}{\textbf{2.51}}                        & \multicolumn{1}{c|}{-}                            & \multicolumn{1}{c|}{-}                           & \multicolumn{1}{c|}{-}                           & \multicolumn{1}{c|}{-}                                    & \multicolumn{1}{c|}{-}                            & \multicolumn{1}{c|}{-}                           & \multicolumn{1}{c|}{-}                                    & -                                           \\ \hline
\multicolumn{1}{|l|}{ViT~\cite{dosovitskiy2020image}}                               & \multicolumn{1}{c|}{7.74}                                 & \multicolumn{1}{c|}{14.96}                        & \multicolumn{1}{c|}{1.85}                        & \multicolumn{1}{c|}{2.57}                        & \multicolumn{1}{c|}{\textbf{0.00}}                        & \multicolumn{1}{c|}{45.63}                        & \multicolumn{1}{c|}{1.19}                        & \multicolumn{1}{c|}{1.98}                                 & 9.74+-16.61                                 \\ \hline
\multicolumn{1}{|l|}{\textbf{ViT+ConvA (Ours)}}         & \multicolumn{1}{c|}{{\color[HTML]{343434} 4.35}}          & \multicolumn{1}{c|}{\textbf{9.67}}                & \multicolumn{1}{c|}{\textbf{0.00}}               & \multicolumn{1}{c|}{\textbf{1.30}}               & \multicolumn{1}{c|}{0.51}                                 & \multicolumn{1}{c|}{45.63}                        & \multicolumn{1}{c|}{0.70}                        & \multicolumn{1}{c|}{0.43}                                 & 8.32+-16.80                                 \\ \hline
\multicolumn{1}{|l|}{\textbf{ViT+ConvA+M$^{2}$A$^{2}$E (Ours)}}     & \multicolumn{1}{c|}{3.34}                                 & \multicolumn{1}{c|}{13.24}                        & \multicolumn{1}{c|}{0.14}                        & \multicolumn{1}{c|}{5.12}                        & \multicolumn{1}{c|}{0.29}                                 & \multicolumn{1}{c|}{\textbf{31.24}}               & \multicolumn{1}{c|}{\textbf{0.61}}               & \multicolumn{1}{c|}{\textbf{0.00}}                        & \textbf{7.23+-11.63}                        \\ \hline
\multicolumn{10}{|c|}{\cellcolor[HTML]{EFEFEF}Modality: \textbf{Depth}}                                                                                                                                        \\ \hline
\multicolumn{1}{|l|}{MC-CNN~\cite{george2019biometric}}                            & \multicolumn{1}{c|}{6.04}                                 & \multicolumn{1}{c|}{-}                            & \multicolumn{1}{c|}{-}                           & \multicolumn{1}{c|}{-}                           & \multicolumn{1}{c|}{-}                                    & \multicolumn{1}{c|}{-}                            & \multicolumn{1}{c|}{-}                           & \multicolumn{1}{c|}{-}                                    & -                                           \\ \hline
\multicolumn{1}{|l|}{ViT~\cite{dosovitskiy2020image}}                               & \multicolumn{1}{c|}{{\color[HTML]{000000} 7.78}}          & \multicolumn{1}{c|}{21.04}                        & \multicolumn{1}{c|}{0.14}                        & \multicolumn{1}{c|}{2.86}                        & \multicolumn{1}{c|}{0.87}                                 & \multicolumn{1}{c|}{37.92}                        & \multicolumn{1}{c|}{1.05}                        & \multicolumn{1}{c|}{9.06}                                 & 10.42+-14.22                                \\ \hline
\multicolumn{1}{|l|}{\textbf{ViT+ConvA (Ours)}}         & \multicolumn{1}{c|}{5.73}                                 & \multicolumn{1}{c|}{26.33}                        & \multicolumn{1}{c|}{0.29}                        & \multicolumn{1}{c|}{\textbf{2.57}}               & \multicolumn{1}{c|}{0.29}                                 & \multicolumn{1}{c|}{36.73}                        & \multicolumn{1}{c|}{0.94}                        & \multicolumn{1}{c|}{\textbf{6.25}}                        & 10.49+-14.83                                \\ \hline
\multicolumn{1}{|l|}{\textbf{ViT+ConvA+M$^{2}$A$^{2}$E (Ours)}}     & \multicolumn{1}{c|}{\textbf{5.31}}                        & \multicolumn{1}{c|}{\textbf{20.27}}               & \multicolumn{1}{c|}{\textbf{0.00}}               & \multicolumn{1}{c|}{\textbf{2.57}}               & \multicolumn{1}{c|}{\textbf{0.00}}                        & \multicolumn{1}{c|}{\textbf{36.55}}               & \multicolumn{1}{c|}{\textbf{0.43}}               & \multicolumn{1}{c|}{7.31}                                 & \textbf{9.59+-13.92}                        \\ \hline
\multicolumn{10}{|c|}{\cellcolor[HTML]{EFEFEF}Modality: \textbf{RGB+IR}}                                                                                                                                           \\ \hline
\multicolumn{1}{|l|}{MA-Net~\cite{liu2021face}}                            & \multicolumn{1}{c|}{6.85}                                 & \multicolumn{1}{c|}{25.33}                        & \multicolumn{1}{c|}{3.16}                        & \multicolumn{1}{c|}{2.05}                        & \multicolumn{1}{c|}{0.28}                                 & \multicolumn{1}{c|}{36.72}                        & \multicolumn{1}{c|}{\textbf{0.86}}               & \multicolumn{1}{c|}{9.82}                                 & 11.18±14.30                                 \\ \hline
\multicolumn{1}{|l|}{ViT~\cite{dosovitskiy2020image}}                               & \multicolumn{1}{c|}{4.02}                                 & \multicolumn{1}{c|}{15.76}                        & \multicolumn{1}{c|}{19.15}                       & \multicolumn{1}{c|}{6.42}                        & \multicolumn{1}{c|}{1.45}                                 & \multicolumn{1}{c|}{\textbf{23.25}}               & \multicolumn{1}{c|}{2.19}                        & \multicolumn{1}{c|}{3.44}                                 & 10.23+-8.96                                 \\ \hline
\multicolumn{1}{|l|}{\textbf{ViT+AMA (Ours)}}               & \multicolumn{1}{c|}{\textbf{1.27}}                        & \multicolumn{1}{c|}{15.49}                        & \multicolumn{1}{c|}{1.16}                        & \multicolumn{1}{c|}{1.74}                        & \multicolumn{1}{c|}{0.43}                                 & \multicolumn{1}{c|}{28.16}                        & \multicolumn{1}{c|}{1.01}                        & \multicolumn{1}{c|}{\textbf{0.77}}                        & 6.97+-11.09                                 \\ \hline
\multicolumn{1}{|l|}{\textbf{ViT+AMA+M$^{2}$A$^{2}$E (Ours)}}           & \multicolumn{1}{c|}{1.35}                                 & \multicolumn{1}{c|}{\textbf{8.63}}                & \multicolumn{1}{c|}{\textbf{0.29}}               & \multicolumn{1}{c|}{\textbf{0.43}}               & \multicolumn{1}{c|}{\textbf{0.00}}                        & \multicolumn{1}{c|}{29.47}                        & \multicolumn{1}{c|}{2.75}                        & \multicolumn{1}{c|}{0.97}                                 & {\color[HTML]{000000} \textbf{6.08+-10.75}} \\ \hline
\multicolumn{10}{|c|}{\cellcolor[HTML]{EFEFEF}Modality: \textbf{RGB+Depth}}                                              
\\ \hline
\multicolumn{1}{|l|}{MC-PixBiS~\cite{george2019deep}}                         & \multicolumn{1}{c|}{1.80}                                 & \multicolumn{1}{c|}{49.70}                        & \multicolumn{1}{c|}{3.70}                        & \multicolumn{1}{c|}{0.70}                        & \multicolumn{1}{c|}{0.10}                                 & \multicolumn{1}{c|}{16.00}                        & \multicolumn{1}{c|}{0.20}                        & \multicolumn{1}{c|}{3.40}                                 & 10.50±16.70                                                               \\ \hline
\multicolumn{1}{|l|}{CMFL~\cite{george2021cross}}                              & \multicolumn{1}{c|}{1.70}                                 & \multicolumn{1}{c|}{12.40}                        & \multicolumn{1}{c|}{1.00}                        & \multicolumn{1}{c|}{2.50}                        & \multicolumn{1}{c|}{0.70}                                 & \multicolumn{1}{c|}{33.50}                        & \multicolumn{1}{c|}{1.80}                        & \multicolumn{1}{c|}{1.70}                                 & 7.60±11.20                                  \\ \hline
\multicolumn{1}{|l|}{MA-ViT~\cite{liuma}}                            & \multicolumn{1}{c|}{1.45}                                 & \multicolumn{1}{c|}{\textbf{9.76}}                & \multicolumn{1}{c|}{0.93}                        & \multicolumn{1}{c|}{0.55}                        & \multicolumn{1}{c|}{\textbf{0.00}}                        & \multicolumn{1}{c|}{\textbf{14.00}}               & \multicolumn{1}{c|}{\textbf{0.00}}               & \multicolumn{1}{c|}{1.46}                                 & \textbf{3.81±5.67}                          \\ \hline
\multicolumn{1}{|l|}{ViT~\cite{dosovitskiy2020image}}                               & \multicolumn{1}{c|}{3.10}                                 & \multicolumn{1}{c|}{18.67}                        & \multicolumn{1}{c|}{8.87}                        & \multicolumn{1}{c|}{5.00}                        & \multicolumn{1}{c|}{0.72}                                 & \multicolumn{1}{c|}{22.52}                        & \multicolumn{1}{c|}{0.58}                        & \multicolumn{1}{c|}{3.93}                                 & 8.61+-8.72                                  \\ \hline
\multicolumn{1}{|l|}{\textbf{ViT+AMA (Ours)}}           & \multicolumn{1}{c|}{{\color[HTML]{343434} \textbf{1.19}}} & \multicolumn{1}{c|}{{\color[HTML]{343434} 18.67}} & \multicolumn{1}{c|}{{\color[HTML]{343434} 1.01}} & \multicolumn{1}{c|}{{\color[HTML]{343434} 2.03}} & \multicolumn{1}{c|}{{\color[HTML]{343434} \textbf{0.00}}} & \multicolumn{1}{c|}{{\color[HTML]{343434} 16.88}} & \multicolumn{1}{c|}{{\color[HTML]{343434} 1.16}} & \multicolumn{1}{c|}{{\color[HTML]{343434} \textbf{0.72}}} & {\color[HTML]{343434} 5.78+-8.23}           \\ \hline
\multicolumn{1}{|l|}{\textbf{ViT+AMA+M$^{2}$A$^{2}$E (Ours)}}       & \multicolumn{1}{c|}{2.53}                                 & \multicolumn{1}{c|}{14.45}                        & \multicolumn{1}{c|}{\textbf{0.29}}               & \multicolumn{1}{c|}{\textbf{0.00}}               & \multicolumn{1}{c|}{\textbf{0.00}}                        & \multicolumn{1}{c|}{19.91}                        & \multicolumn{1}{c|}{3.91}                        & \multicolumn{1}{c|}{2.19}                                 & 5.82+-8.04                                  \\ \hline
\multicolumn{10}{|c|}{\cellcolor[HTML]{EFEFEF}Modality: \textbf{IR+Depth}}                                                                                                                                                                                           \\ \hline
\multicolumn{1}{|l|}{ViT~\cite{dosovitskiy2020image}}                               & \multicolumn{1}{c|}{5.47}                                 & \multicolumn{1}{c|}{15.64}                        & \multicolumn{1}{c|}{\textbf{0.00}}               & \multicolumn{1}{c|}{1.71}                        & \multicolumn{1}{c|}{\textbf{0.00}}                        & \multicolumn{1}{c|}{41.03}                        & \multicolumn{1}{c|}{0.67}                        & \multicolumn{1}{c|}{2.38}                                 & 8.78+-15.26                                 \\ \hline
\multicolumn{1}{|l|}{\textbf{ViT+AMA (Ours)}}           & \multicolumn{1}{c|}{\textbf{1.88}}                        & \multicolumn{1}{c|}{\textbf{10.45}}               & \multicolumn{1}{c|}{\textbf{0.00}}               & \multicolumn{1}{c|}{3.92}                        & \multicolumn{1}{c|}{\textbf{0.00}}                        & \multicolumn{1}{c|}{41}                           & \multicolumn{1}{c|}{0.38}                        & \multicolumn{1}{c|}{1.55}                                 & 8.19+-14.94                                 \\ \hline
\multicolumn{1}{|l|}{\textbf{ViT+AMA+M$^{2}$A$^{2}$E (Ours)}}       & \multicolumn{1}{c|}{1.96}                                 & \multicolumn{1}{c|}{11.51}                        & \multicolumn{1}{c|}{\textbf{0.00}}               & \multicolumn{1}{c|}{\textbf{0.98}}               & \multicolumn{1}{c|}{\textbf{0.00}}                        & \multicolumn{1}{c|}{\textbf{34.06}}               & \multicolumn{1}{c|}{\textbf{0.00}}               & \multicolumn{1}{c|}{\textbf{0.00}}                        & \textbf{6.65+-12.81}                        \\ \hline
\multicolumn{10}{|c|}{\cellcolor[HTML]{EFEFEF}Modality: \textbf{RGB+IR+Depth} \quad (* indicates with RGB+IR+Depth+Thermal modalities)}                                                                                               \\ \hline
\multicolumn{1}{|l|}{IQM+LBP*~\cite{george2019biometric}}                          & \multicolumn{1}{c|}{7.54}                                 & \multicolumn{1}{c|}{28.58}                        & \multicolumn{1}{c|}{0.84}                        & \multicolumn{1}{c|}{2.38}                        & \multicolumn{1}{c|}{2.3}                                  & \multicolumn{1}{c|}{50.86}                        & \multicolumn{1}{c|}{16.34}                       & \multicolumn{1}{c|}{14.27}                                & 16.29±18.36                                 \\ \hline
\multicolumn{1}{|l|}{MC-CNN*~\cite{george2019biometric}}                           & \multicolumn{1}{c|}{1.04}                                 & \multicolumn{1}{c|}{\textbf{2.52}}                & \multicolumn{1}{c|}{0.12}                        & \multicolumn{1}{c|}{\textbf{0.00}}               & \multicolumn{1}{c|}{\textbf{0.00}}                        & \multicolumn{1}{c|}{42.14}                        & \multicolumn{1}{c|}{0.35}                        & \multicolumn{1}{c|}{0.75}                                 & 6.55±15.72                                  \\ \hline
\multicolumn{1}{|l|}{ViT~\cite{dosovitskiy2020image}}                               & \multicolumn{1}{c|}{{\color[HTML]{000000} 2.52}}          & \multicolumn{1}{c|}{15.88}                        & \multicolumn{1}{c|}{5.46}                        & \multicolumn{1}{c|}{0.58}                        & \multicolumn{1}{c|}{0.14}                                 & \multicolumn{1}{c|}{19.99}                        & \multicolumn{1}{c|}{0.90}                        & \multicolumn{1}{c|}{2.32}                                 & 6.46+-8.12                                  \\ \hline
\multicolumn{1}{|l|}{\textbf{ViT+AMA (Ours)}}           & \multicolumn{1}{c|}{\textbf{0.92}}                        & \multicolumn{1}{c|}{15.39}                        & \multicolumn{1}{c|}{0.64}                        & \multicolumn{1}{c|}{1.99}                        & \multicolumn{1}{c|}{0.87}                                 & \multicolumn{1}{c|}{18.37}                        & \multicolumn{1}{c|}{0.87}                        & \multicolumn{1}{c|}{0.77}                                 & {\color[HTML]{000000} 5.56+-7.80}           \\ \hline
\multicolumn{1}{|l|}{\textbf{ViT+AMA+M$^{2}$A$^{2}$E (Ours)}}       & \multicolumn{1}{c|}{1.39}                                 & \multicolumn{1}{c|}{9.02}                         & \multicolumn{1}{c|}{\textbf{0.00}}               & \multicolumn{1}{c|}{\textbf{0.00}}               & \multicolumn{1}{c|}{\textbf{0.00}}                        & \multicolumn{1}{c|}{\textbf{17.99}}               & \multicolumn{1}{c|}{\textbf{0.00}}               & \multicolumn{1}{c|}{\textbf{0.00}}                        & {\color[HTML]{000000} \textbf{3.86+-7.08}}  \\ \hline
\end{tabular}}
\vspace{-1.4em}
\end{table*}

As shown in Fig.~\ref{fig:MAE}, given a multimodal face sample ($X_{\text{RGB}},X_{\text{IR}},X_{\text{Depth}}$), M$^{2}$A$^{2}$E randomly selects unimodal input $X_{i}$ ($i\in{\text{RGB},\text{IR},\text{Depth}}$) among all modalities. Then random sampling strategy~\cite{he2022masked} is used to mask out $p$ percentage of the visual tokens in $X_{i}$. Only the unmasked visible tokens are forwarded the ViT encoder and both visible and masked tokens are fed in unshared ViT decoders. In terms of the reconstruction target, given a masked input $X_{i}$ with the $i$-th modality, M$^{2}$A$^{2}$E aims to predict the pixel values with mean squared error (MSE) loss for 1) each masked patch of $X_{i}$, and 2) the whole input images of other modalities $X_{j}$ ($j\neq i; j\in{\text{RGB},\text{IR},\text{Depth}}$). The motivation behind M$^{2}$A$^{2}$E is that with the multimodal reconstruction target, the self-supervised pre-trained ViTs are able to model 1) task-aware contextual semantics (e.g., moiré patterns and color distortion) via masked patch prediction; and 2) intrinsic physical features (e.g., 2D attacks without facial depth) via cross modality translation.

\vspace{0.2em}
\noindent\textbf{Relation to modality-symmetric autoencoders~\cite{he2022masked,bachmann2022multimae}.}\quad    Compared with the vanilla MAE~\cite{he2022masked}, M$^{2}$A$^{2}$E adopts the same masked strategy in unimodal ViT encoder but targeting at multimodal reconstruction with multiple unshared ViT decoders. Besides, M$^{2}$A$^{2}$E is similar to the multimodal MAE~\cite{bachmann2022multimae} only when partial tokens from a single modality are visible while masking all tokens from other modalities.

\vspace{-0.3em}
\section{Experimental Evaluation}
\vspace{-0.2em}
\label{sec:experiemnts}

\subsection{Datasets and Performance Metrics}
\label{sec:dataset} 
\vspace{-0.2em}

Three commonly used multimodal FAS datasets are used for experiments, including WMCA~\cite{george2019biometric}, CASIA-SURF (MmFA)~\cite{zhang2019dataset} and CASIA-SURF CeFA (CeFA)~\cite{liu2021casia}. \textbf{WMCA} contains a wide variety of 2D and 3D PAs with four modalities, which introduces 2 protocols: `seen' protocol which emulates the seen attack scenario and the `unseen' attack protocol that evaluates the generalization on an unseen attack. \textbf{MmFA} consists of 1000 subjects with 21000 videos, and each sample has 3 modalities, which has an official intra-testing protocol. \textbf{CeFA} is the largest multimodal FAS dataset, covering 3 ethnicities, 3 modalities, 1607 subjects, and 34200 videos. We conduct intra- and cross-dataset testings on WMCA and MmFA datasets, and leave large-scale CeFA for self-supervised pre-training.  

In terms of evaluation metrics, Attack Presentation Classification Error Rate (APCER), Bonafide Presentation Classification Error Rate (BPCER), and ACER~\cite{ACER} are used for the metrics. The ACER on testing set is determined by the Equal Error Rate (EER) threshold on dev sets for MmFA, and the BPCER=1\% threshold for WMCA. True Positive Rate (TPR)@False Positive Rate (FPR)=10$^{-4}$~\cite{zhang2019dataset} is also provided for MmFA. For cross-testing experiments, Half Total Error Rate (HTER) is adopted. 

\subsection{Implementation Details}
\label{sec:Details}

We crop the face frames using MTCNN~\cite{zhang2016joint} face detector. The local descriptors are extracted from gray-scale images with: 1) 3x3 neighbors for LBP~\cite{ojala2002multiresolution}; 2) 9 orientations, 8x8 pixels per cell, and 2x2 cells per block for HOG~\cite{dalal2005histograms}; and 3) the size of masks are set to 5 for PLGF~\cite{bhattacharjee2019pattern}. Composition inputs `GRAY\_HOG\_PLGF' are adopted on unimodal and multimodal experiments for IR modality, while the raw inputs are utilized for RGB and Depth modalities. ViT-Base~\cite{dosovitskiy2020image} supervised by binary cross-entropy loss is used as the defaulted architecture. For the direct finetuning, only the last transformer block and classification head are trainable. For AMA and ConvAdapter~\cite{jie2022convolutional} finetuning, the original and hidden channels are $D$=768 and $D'$=64, respectively. For M$^{2}$A$^{2}$E, the mask ratio $p$=40\% is used while decoder depth and width is 4 and 512, respectively.

The experiments are implemented with Pytorch on one NVIDIA A100 GPU. For the self-supervised pre-training on CeFA with RGB+IR+Depth modalities, we use the AdamW~\cite{loshchilov2017decoupled} optimizer with learning rate (lr) 1.5e-4, weight decay (wd) 0.05 and batch size 64 at the training stage. ImageNet pre-trained weights are used for our encoder. We train the M$^{2}$A$^{2}$E for 400 epochs while warming up the first 40 epochs, and then performing cosine decay. For supervised unimodal and multimodal experiments on WMCA and MmFA, we use the Adam optimizer with the fixed lr=2e-4, wd=5e-3 and batch size 16 at the training stage. We finetune models with maximum 30 epochs based on the ImageNet or M$^{2}$A$^{2}$E pre-trained weights.

\begin{table}[t]
\centering
\caption{ The results on MmFA. Larger TPR and lower ACER values indicate better performance. Best results are marked in \textbf{bold}.} \label{tab:MmFA}

\vspace{-0.8em}

\resizebox{0.48\textwidth}{!}{\begin{tabular}{|lcccc|}
\hline
\multicolumn{1}{|c|}{\textbf{Method}}               & \multicolumn{1}{c|}{\textbf{APCER(\%)}}               & \multicolumn{1}{c|}{\textbf{BPCER(\%)}} & \multicolumn{1}{c|}{\textbf{ACER(\%)}} & \textbf{\begin{tabular}[c]{@{}c@{}}TPR(\%)\\ @FPR=10$^{-4}$\end{tabular}} \\ \hline
\multicolumn{5}{|c|}{\cellcolor[HTML]{EFEFEF}Modality: \textbf{RGB}}                                                                                                                                                       \\ \hline
\multicolumn{1}{|l|}{SEF~\cite{zhang2019dataset}}                           & \multicolumn{1}{c|}{8.0}                              & \multicolumn{1}{c|}{14.5}               & \multicolumn{1}{c|}{11.3}              & 6.8                       \\ \hline
\multicolumn{1}{|l|}{MS-SEF~\cite{zhang2020casia}}                        & \multicolumn{1}{c|}{40.3}                              & \multicolumn{1}{c|}{1.6}               & \multicolumn{1}{c|}{21.0}              & 14.6                      \\ \hline
\multicolumn{1}{|l|}{ViT~\cite{dosovitskiy2020image}}                           & \multicolumn{1}{c|}{{\color[HTML]{000000} 18.91}} & \multicolumn{1}{c|}{15.83}                   & \multicolumn{1}{c|}{17.37}                  &    16.72                       \\ \hline
\multicolumn{1}{|l|}{\textbf{ViT+ConvA (Ours)}}     & \multicolumn{1}{c|}{{\color[HTML]{000000} 10.70}} & \multicolumn{1}{c|}{9.33}                   & \multicolumn{1}{c|}{10.02}                  &   20.22                        \\ \hline
\multicolumn{1}{|l|}{\textbf{ViT+ConvA+M$^{2}$A$^{2}$E (Ours)}} & \multicolumn{1}{c|}{6.62}                                 & \multicolumn{1}{c|}{6.17}                   & \multicolumn{1}{c|}{\textbf{6.40}}                  &   \textbf{23.77}                        \\ \hline
\multicolumn{5}{|c|}{\cellcolor[HTML]{EFEFEF}Modality: \textbf{IR}}                                                                                                                                                        \\ \hline
\multicolumn{1}{|l|}{SEF~\cite{zhang2019dataset}}                           & \multicolumn{1}{c|}{15.0}                             & \multicolumn{1}{c|}{1.2}                & \multicolumn{1}{c|}{\textbf{8.1}}               & 10.9                      \\ \hline
\multicolumn{1}{|l|}{MS-SEF~\cite{zhang2020casia}}                        & \multicolumn{1}{c|}{38.6}                              & \multicolumn{1}{c|}{0.4}               & \multicolumn{1}{c|}{19.4}              & 15.9                      \\ \hline
\multicolumn{1}{|l|}{ViT~\cite{dosovitskiy2020image}}                           & \multicolumn{1}{c|}{18.74}                        & \multicolumn{1}{c|}{19.11}          & \multicolumn{1}{c|}{18.92}         &   10.72                        \\ \hline
\multicolumn{1}{|l|}{\textbf{ViT+ConvA (Ours)}}     & \multicolumn{1}{c|}{{\color[HTML]{343434} 16.30}} & \multicolumn{1}{c|}{13.00}          & \multicolumn{1}{c|}{14.65}         &     17.36                      \\ \hline
\multicolumn{1}{|l|}{\textbf{ViT+ConvA+M$^{2}$A$^{2}$E (Ours)}} & \multicolumn{1}{c|}{12.73}                                 & \multicolumn{1}{c|}{11.44}                   & \multicolumn{1}{c|}{12.09}     & \textbf{19.94}                \\ \hline
\multicolumn{5}{|c|}{\cellcolor[HTML]{EFEFEF}Modality: \textbf{Depth}}                                                                                                                                                     \\ \hline
\multicolumn{1}{|l|}{SEF~\cite{zhang2019dataset}}                           & \multicolumn{1}{c|}{5.1}                              & \multicolumn{1}{c|}{4.8}                & \multicolumn{1}{c|}{5.0}               & 14.1                      \\ \hline
\multicolumn{1}{|l|}{MS-SEF~\cite{zhang2020casia}}                        & \multicolumn{1}{c|}{6.0}                              & \multicolumn{1}{c|}{1.2}                & \multicolumn{1}{c|}{3.6}               & \textbf{67.3}                      \\ \hline
\multicolumn{1}{|l|}{ViT~\cite{dosovitskiy2020image}}                           & \multicolumn{1}{c|}{{\color[HTML]{000000} 2.67 }}          & \multicolumn{1}{c|}{2.22}          & \multicolumn{1}{c|}{2.44}         &   36.56                        \\ \hline
\multicolumn{1}{|l|}{\textbf{ViT+ConvA (Ours)}}     & \multicolumn{1}{c|}{1.19}                        & \multicolumn{1}{c|}{2.61}                   & \multicolumn{1}{c|}{\textbf{1.90}}         &       64.11                    \\ \hline
\multicolumn{1}{|l|}{\textbf{ViT+ConvA+M$^{2}$A$^{2}$E (Ours)}} & \multicolumn{1}{c|}{1.68}                        & \multicolumn{1}{c|}{2.61}          & \multicolumn{1}{c|}{2.15}         & 51.39                \\ \hline
\multicolumn{5}{|c|}{\cellcolor[HTML]{EFEFEF}Modality: \textbf{RGB+IR}}                                                                                                                                                    \\ \hline
\multicolumn{1}{|l|}{SEF~\cite{zhang2019dataset}}                           & \multicolumn{1}{c|}{14.4}                             & \multicolumn{1}{c|}{1.6}                & \multicolumn{1}{c|}{8.0}               & 26.1                      \\ \hline
\multicolumn{1}{|l|}{MS-SEF~\cite{zhang2020casia}}                        & \multicolumn{1}{c|}{36.5}                             & \multicolumn{1}{c|}{0.005}              & \multicolumn{1}{c|}{18.3}              & 37.0                      \\ \hline
\multicolumn{1}{|l|}{ViT~\cite{dosovitskiy2020image}}                           & \multicolumn{1}{c|}{17.18}                                 & \multicolumn{1}{c|}{18.94}          & \multicolumn{1}{c|}{18.06}         &       24.67                    \\ \hline
\multicolumn{1}{|l|}{\textbf{ViT+AMA (Ours)}}     & \multicolumn{1}{c|}{{\color[HTML]{343434} 16.83}} & \multicolumn{1}{c|}{11.87}          & \multicolumn{1}{c|}{14.35}         &      36.67                     \\ \hline
\multicolumn{1}{|l|}{\textbf{ViT+AMA+M$^{2}$A$^{2}$E (Ours)}} & \multicolumn{1}{c|}{11.38}                                 & \multicolumn{1}{c|}{11.44}          & \multicolumn{1}{c|}{\textbf{11.41}}         & \textbf{40.94}                 \\ \hline
\multicolumn{5}{|c|}{\cellcolor[HTML]{EFEFEF}Modality: \textbf{RGB+Depth}}                                                                                                                                                 \\ \hline
\multicolumn{1}{|l|}{SEF~\cite{zhang2019dataset}}                           & \multicolumn{1}{c|}{4.3}                              & \multicolumn{1}{c|}{5.6}                & \multicolumn{1}{c|}{5.0}               & 10.6                      \\ \hline
\multicolumn{1}{|l|}{MS-SEF~\cite{zhang2020casia}}                        & \multicolumn{1}{c|}{5.8}                              & \multicolumn{1}{c|}{0.8}                & \multicolumn{1}{c|}{3.3}               & 71.1                      \\ \hline
\multicolumn{1}{|l|}{ViT~\cite{dosovitskiy2020image}}                           & \multicolumn{1}{c|}{5.13}                        & \multicolumn{1}{c|}{4.06}          & \multicolumn{1}{c|}{4.60}         &      36.22                     \\ \hline
\multicolumn{1}{|l|}{\textbf{ViT+AMA (Ours)}}     & \multicolumn{1}{c|}{1.29}                        & \multicolumn{1}{c|}{2.39}                   & \multicolumn{1}{c|}{1.84}         &       67.89                \\ \hline
\multicolumn{1}{|l|}{\textbf{ViT+AMA+M$^{2}$A$^{2}$E (Ours)}} & \multicolumn{1}{c|}{1.25}                        & \multicolumn{1}{c|}{2.06}                   & \multicolumn{1}{c|}{\textbf{1.65}}         &       \textbf{75.06}             \\ \hline
\multicolumn{5}{|c|}{\cellcolor[HTML]{EFEFEF}Modality: \textbf{IR+Depth}}                                                                                                                                                  \\ \hline
\multicolumn{1}{|l|}{SEF~\cite{zhang2019dataset}}                           & \multicolumn{1}{c|}{1.5}                              & \multicolumn{1}{c|}{8.4}                & \multicolumn{1}{c|}{4.9}               & 24.3                      \\ \hline
\multicolumn{1}{|l|}{MS-SEF~\cite{zhang2020casia}}                        & \multicolumn{1}{c|}{2.0}                              & \multicolumn{1}{c|}{0.3}                & \multicolumn{1}{c|}{\textbf{1.1}}               & \textbf{81.2}                      \\ \hline
\multicolumn{1}{|l|}{ViT~\cite{dosovitskiy2020image}}                           & \multicolumn{1}{c|}{2.08}                        & \multicolumn{1}{c|}{3.28}          & \multicolumn{1}{c|}{2.68}         &     40.39                      \\ \hline
\multicolumn{1}{|l|}{\textbf{ViT+AMA (Ours)}}     & \multicolumn{1}{c|}{{\color[HTML]{343434} 1.56}} & \multicolumn{1}{c|}{1.78}          & \multicolumn{1}{c|}{1.67}         & 59.72                 \\ \hline
\multicolumn{1}{|l|}{\textbf{ViT+AMA+M$^{2}$A$^{2}$E (Ours)}} & \multicolumn{1}{c|}{1.48}                                 & \multicolumn{1}{c|}{0.83}          & \multicolumn{1}{c|}{1.16}         & 67.33                 \\ \hline
\multicolumn{5}{|c|}{\cellcolor[HTML]{EFEFEF}Modality: \textbf{RGB+IR+Depth}}                                                                                                                                           \\ \hline
\multicolumn{1}{|l|}{SEF~\cite{zhang2019dataset}}                           & \multicolumn{1}{c|}{3.8}                             & \multicolumn{1}{c|}{1.0}               & \multicolumn{1}{c|}{2.4}              & 56.8                     \\ \hline
\multicolumn{1}{|l|}{MS-SEF~\cite{zhang2020casia}}                        & \multicolumn{1}{c|}{1.9}                             & \multicolumn{1}{c|}{0.1}               & \multicolumn{1}{c|}{1.0}              & \textbf{92.4}                     \\ \hline
\multicolumn{1}{|l|}{MA-ViT~\cite{liuma}}                        & \multicolumn{1}{c|}{0.78}                             & \multicolumn{1}{c|}{0.83}               & \multicolumn{1}{c|}{0.80}              & 82.83                     \\ \hline
\multicolumn{1}{|l|}{ViT~\cite{dosovitskiy2020image}}                           & \multicolumn{1}{c|}{{\color[HTML]{000000} 2.10}}          & \multicolumn{1}{c|}{1.78}          & \multicolumn{1}{c|}{1.94}         &    66.61                       \\ \hline
\multicolumn{1}{|l|}{\textbf{ViT+AMA (Ours)}}     & \multicolumn{1}{c|}{{\color[HTML]{343434} 2.22}} & \multicolumn{1}{c|}{0.49}          & \multicolumn{1}{c|}{1.36}         & 78.94                 \\ \hline
\multicolumn{1}{|l|}{\textbf{ViT+AMA+M$^{2}$A$^{2}$E (Ours)}} & \multicolumn{1}{c|}{0.81}                                 & \multicolumn{1}{c|}{0.42}          & \multicolumn{1}{c|}{\textbf{0.62}}         & 85.23               \\ \hline
\end{tabular}}
\vspace{-1.4em}
\end{table}

\begin{table}[t]
\centering
\caption{The HTER (\%) values from the cross-testing between WMCA and MmFA datasets. Best results are marked in \textbf{bold}.} \label{tab:Crosstesting}

\vspace{-0.8em}

\resizebox{0.42\textwidth}{!}{
\begin{tabular}{|lclcl|}
\hline
\multicolumn{1}{|c|}{\textbf{Method}}               & \multicolumn{2}{c|}{\textbf{\begin{tabular}[c]{@{}c@{}}Train on WMCA\\ Test on MmFA\end{tabular}}} & \multicolumn{2}{c|}{\textbf{\begin{tabular}[c]{@{}c@{}}Train on MmFA\\ Test on WMCA\end{tabular}}} \\ \hline
\multicolumn{5}{|c|}{\cellcolor[HTML]{EFEFEF}Modality: \textbf{RGB}}                                                                                                                                                                                          \\ \hline
\multicolumn{1}{|l|}{CDCN~\cite{yu2020searching}}                          & \multicolumn{2}{c|}{27.47}                                                                              & \multicolumn{2}{c|}{34.66}                                                                              \\ \hline
\multicolumn{1}{|l|}{ViT~\cite{dosovitskiy2020image}}                           & \multicolumn{2}{c|}{{\color[HTML]{000000} 25.18}}                                              & \multicolumn{2}{c|}{50.26}                                                                              \\ \hline
\multicolumn{1}{|l|}{\textbf{ViT+ConvA (Ours)}}     & \multicolumn{2}{c|}{{\color[HTML]{000000} \textbf{21.31}}}                                              & \multicolumn{2}{c|}{\textbf{35.29}}                                                                              \\ \hline
\multicolumn{1}{|l|}{\textbf{ViT+ConvA+M$^{2}$A$^{2}$E (Ours)}} & \multicolumn{2}{c|}{23.56}                                                                              & \multicolumn{2}{c|}{35.69}                                                                              \\ \hline
\multicolumn{5}{|c|}{\cellcolor[HTML]{EFEFEF}Modality: \textbf{IR}}                                                                                                                                                                                           \\ \hline
\multicolumn{1}{|l|}{CDCN~\cite{yu2020searching}}                          & \multicolumn{2}{c|}{44.11}                                                                     & \multicolumn{2}{c|}{31.19}                                                                     \\ \hline
\multicolumn{1}{|l|}{ViT~\cite{dosovitskiy2020image}}                           & \multicolumn{2}{c|}{35.22}                                                                              & \multicolumn{2}{c|}{37.88}                                                                              \\ \hline
\multicolumn{1}{|l|}{\textbf{ViT+ConvA (Ours)}}     & \multicolumn{2}{c|}{{\color[HTML]{343434} 30.61}}                                              & \multicolumn{2}{c|}{31.78}                                                                     \\ \hline
\multicolumn{1}{|l|}{\textbf{ViT+ConvA+M$^{2}$A$^{2}$E (Ours)}} & \multicolumn{2}{c|}{\textbf{26.06}}                                                                              & \multicolumn{2}{c|}{\textbf{26.50}}                                                                              \\ \hline
\multicolumn{5}{|c|}{\cellcolor[HTML]{EFEFEF}Modality: \textbf{Depth}}                                                                                                                                                                                        \\ \hline
\multicolumn{1}{|l|}{CDCN~\cite{yu2020searching}}                          & \multicolumn{2}{c|}{31.16}                                                                              & \multicolumn{2}{c|}{32.11}                                                                              \\ \hline
\multicolumn{1}{|l|}{ViT~\cite{dosovitskiy2020image}}                           & \multicolumn{2}{c|}{{\color[HTML]{000000} 27.94}}                                                       & \multicolumn{2}{c|}{29.53}                                                                     \\ \hline
\multicolumn{1}{|l|}{\textbf{ViT+ConvA (Ours)}}     & \multicolumn{2}{c|}{25.04}                                                                     & \multicolumn{2}{c|}{26.75}                                                                     \\ \hline
\multicolumn{1}{|l|}{\textbf{ViT+ConvA+M$^{2}$A$^{2}$E (Ours)}} & \multicolumn{2}{c|}{\textbf{23.12}}                                                                     & \multicolumn{2}{c|}{\textbf{23.71}}                                                                     \\ \hline
\multicolumn{5}{|c|}{\cellcolor[HTML]{EFEFEF}Modality: \textbf{RGB+IR}}                                                                                                                                                                                       \\ \hline
\multicolumn{1}{|l|}{MM-CDCN~\cite{yu2020multi}}                       & \multicolumn{2}{c|}{24.60}                                                                              & \multicolumn{2}{c|}{27.86}                                                                              \\ \hline
\multicolumn{1}{|l|}{ViT~\cite{dosovitskiy2020image}}                           & \multicolumn{2}{c|}{27.07}                                                                              & \multicolumn{2}{c|}{30.63}                                                                     \\ \hline
\multicolumn{1}{|l|}{\textbf{ViT+AMA (Ours)}}     & \multicolumn{2}{c|}{{\color[HTML]{343434} 19.77}}                                              & \multicolumn{2}{c|}{27.80}                                                                     \\ \hline
\multicolumn{1}{|l|}{\textbf{ViT+AMA+M$^{2}$A$^{2}$E (Ours)}} & \multicolumn{2}{c|}{\textbf{17.82}}                                                                              & \multicolumn{2}{c|}{\textbf{25.67}}                                                                     \\ \hline
\multicolumn{5}{|c|}{\cellcolor[HTML]{EFEFEF}Modality: \textbf{RGB+Depth}}                                                                                                                                                                                    \\ \hline
\multicolumn{1}{|l|}{MM-CDCN~\cite{yu2020multi}}                       & \multicolumn{2}{c|}{22.38}                                                                              & \multicolumn{2}{c|}{25.46}                                                                              \\ \hline
\multicolumn{1}{|l|}{ViT~\cite{dosovitskiy2020image}}                           & \multicolumn{2}{c|}{23.72}                                                                     & \multicolumn{2}{c|}{29.95}                                                                     \\ \hline
\multicolumn{1}{|l|}{\textbf{ViT+AMA (Ours)}}     & \multicolumn{2}{c|}{{\color[HTML]{343434} 19.17}}                                              & \multicolumn{2}{c|}{\textbf{24.99}}                                                                     \\ \hline
\multicolumn{1}{|l|}{\textbf{ViT+AMA+M$^{2}$A$^{2}$E (Ours)}} & \multicolumn{2}{c|}{\textbf{18.34}}                                                                              & \multicolumn{2}{c|}{25.82}                                                                     \\ \hline
\multicolumn{5}{|c|}{\cellcolor[HTML]{EFEFEF}Modality: \textbf{IR+Depth}}                                                                                                                                                                                     \\ \hline
\multicolumn{1}{|l|}{MM-CDCN~\cite{yu2020multi}}                       & \multicolumn{2}{c|}{29.5}                                                                              & \multicolumn{2}{c|}{30.22}                                                                              \\ \hline
\multicolumn{1}{|l|}{ViT~\cite{dosovitskiy2020image}}                           & \multicolumn{2}{c|}{31.84}                                                                     & \multicolumn{2}{c|}{34.99}                                                                     \\ \hline
\multicolumn{1}{|l|}{\textbf{ViT+AMA (Ours)}}     & \multicolumn{2}{c|}{{\color[HTML]{343434} 25.49}}                                              & \multicolumn{2}{c|}{28.04}                                                                     \\ \hline
\multicolumn{1}{|l|}{\textbf{ViT+AMA+M$^{2}$A$^{2}$E (Ours)}} & \multicolumn{2}{c|}{\textbf{21.43}}                                                                              & \multicolumn{2}{c|}{\textbf{26.35}}                                                                     \\ \hline
\multicolumn{5}{|c|}{\cellcolor[HTML]{EFEFEF}Modality: \textbf{RGB+IR+Depth}}                                                                                                                                                                                 \\ \hline
\multicolumn{1}{|l|}{Aux.(Depth)~\cite{Liu2018Learning}}                   & \multicolumn{2}{c|}{12.35}                                                                         & \multicolumn{2}{c|}{24.54}                                                                         \\ \hline
\multicolumn{1}{|l|}{MM-CDCN~\cite{yu2020multi}}                       & \multicolumn{2}{c|}{21.25}                                                                              & \multicolumn{2}{c|}{21.83}                                                                              \\ \hline
\multicolumn{1}{|l|}{MA-ViT~\cite{liuma}}                        & \multicolumn{2}{c|}{10.41}                                                                         & \multicolumn{2}{c|}{20.63}                                                                         \\ \hline
\multicolumn{1}{|l|}{ViT~\cite{dosovitskiy2020image}}                           & \multicolumn{2}{c|}{{\color[HTML]{000000} 19.19}}                                                       & \multicolumn{2}{c|}{23.21}                                                                     \\ \hline
\multicolumn{1}{|l|}{\textbf{ViT+AMA (Ours)}}     & \multicolumn{2}{c|}{{\color[HTML]{343434} 13.99}}                                              & \multicolumn{2}{c|}{20.22}                                                                     \\ \hline
\multicolumn{1}{|l|}{\textbf{ViT+AMA+M$^{2}$A$^{2}$E (Ours)}} & \multicolumn{2}{c|}{\textbf{8.60}}                                                                              & \multicolumn{2}{c|}{\textbf{18.83}}                                                                     \\ \hline
\end{tabular}}
\vspace{-1.9em}
\end{table}

\subsection{Intra-dataset Testing}

\noindent\textbf{Intra testing on WMCA.} \quad   
The unimodal and multimodal results of protocols ‘seen’ and ‘unseen’ on WMCA~\cite{george2019biometric} are shown in Table~\ref{tab:WMCA}. On the one hand, compared with the direct finetuning results from `ViT', the ViT+AMA/ConvAdapter can achieve significantly lower ACER in all modalities settings and both `seen' and `unseen' protocols. This indicates the proposed AMA efficiently leverages the unimodal/multimodal local inductive cues to boost original ViT's global contextual features. On the other hand, when replaced the ImageNet pre-trained ViT with self-supervised M$^{2}$A$^{2}$E from CeFA, the generalization for unseen attack detection improves obviously with modalities `IR', `Depth', `RGB+IR', `IR+Depth', and `RGB+IR+Depth', indicating its excellent transferability for downstream modality-agnostic tasks. It is surprising to find in the last block that the proposed methods with RGB+IR+Depth modalities perform even better than `MC-CNN'~\cite{george2019biometric} with four modalities in both `seen' and `unseen' protocols. With complex and specialized modules, although `MA-ViT'~\cite{liuma} outperforms the proposed methods with RGB+Depth modalities in `unseen' protocol by -2.01\% ACER, the proposed AMA and M$^{2}$A$^{2}$E might be potential to plug-and-play in `MA-ViT' for further performance improvement. 

\begin{figure*}
\centering
\vspace{-1.0em}
\includegraphics[scale=0.5]{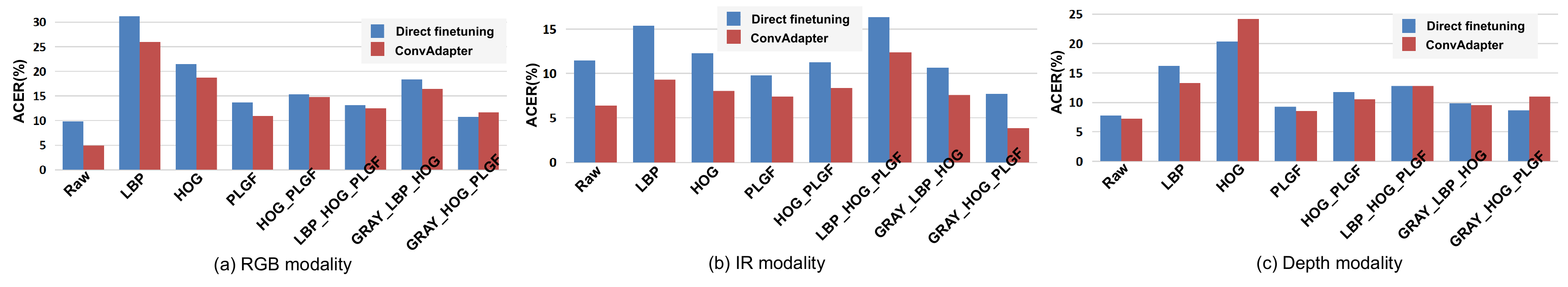}
\vspace{-2.3em}
  \caption{\small{
 Impacts of inputs with local feature descriptors (e.g., LBP, HOG, PLGF) for ViT using direct finetuning and ConvAdapter strategies on (a) RGB, (b) IR, and (c) Depth modalities. More results on multimodal settings can be found in \textit{Appendix A}. }
  }
\label{fig:LBP}
\vspace{-0.8em}
\end{figure*}

\vspace{0.3em}
\noindent\textbf{Intra testing on MmFA.} \quad   
For MmFA, we compare with three famous multimodal methods `SEF'~\cite{zhang2019dataset}, `MS-SEF'~\cite{zhang2020casia}, and `MA-ViT'~\cite{liuma}. From Table~\ref{tab:MmFA}, we can observe that the performance of `ViT' is usually worse than `MS-SEF' with multimodal settings due to the limited modality fusion ability. When equipped with AMA and M$^{2}$A$^{2}$E-based self-supervised pre-training, `ViT+AMA+M$^{2}$A$^{2}$E' outperforms MS-SEF by a large margin in most modality settings (`RGB', `Depth', `RGB+IR', `RGB+Depth', `RGB+IR+Depth') in terms of ACER metrics. Thanks to the powerful multimodal representation capacity from the M$^{2}$A$^{2}$E pre-trained model, the proposed method surpasses the dedicated `MA-ViT' with `RGB+IR+Depth' modalities.

\vspace{-0.3em}
\subsection{Cross-dataset Testing}
\vspace{-0.3em}
To evaluate the unimodal and multimodal generalization, we conduct cross-testing experiments between models trained on MmFA and WMCA with Protocol `seen'. We also introduce the `MM-CDCN'~\cite{yu2020multi} and `MA-ViT'~\cite{liuma} as baselines. Table~\ref{tab:Crosstesting} lists the HTER of all methods trained on one dataset and tested on another dataset. From these results, the proposed `ViT+AMA+M$^{2}$A$^{2}$E' outperforms `MM-CDCN' in most modality settings and `MA-ViT' with `RGB+IR+Depth' on both two cross-testing protocols, indicating that the learned multimodal features are robust to the sensors, resolutions, and attack types. Specifically, directly finetuning ImageNet pre-trained ViTs (see results of `ViT') usually generalize worse than `MM-CDCN' in multimodal settings. When assembling with AMA and M$^{2}$A$^{2}$E, the HTER can be further reduced 9.25\%/5.38\%/10.41\%/10.59\% and 4.96\%/4.13\%/8.64\%/4.38\% for `RGB+IR'/`RGB+Depth'/ `IR+Depth'/`RGB+IR+Depth' when tested on MmFA and WMCA, respectively.

\begin{figure}
\centering
\vspace{-1.0em}
\includegraphics[scale=0.48]{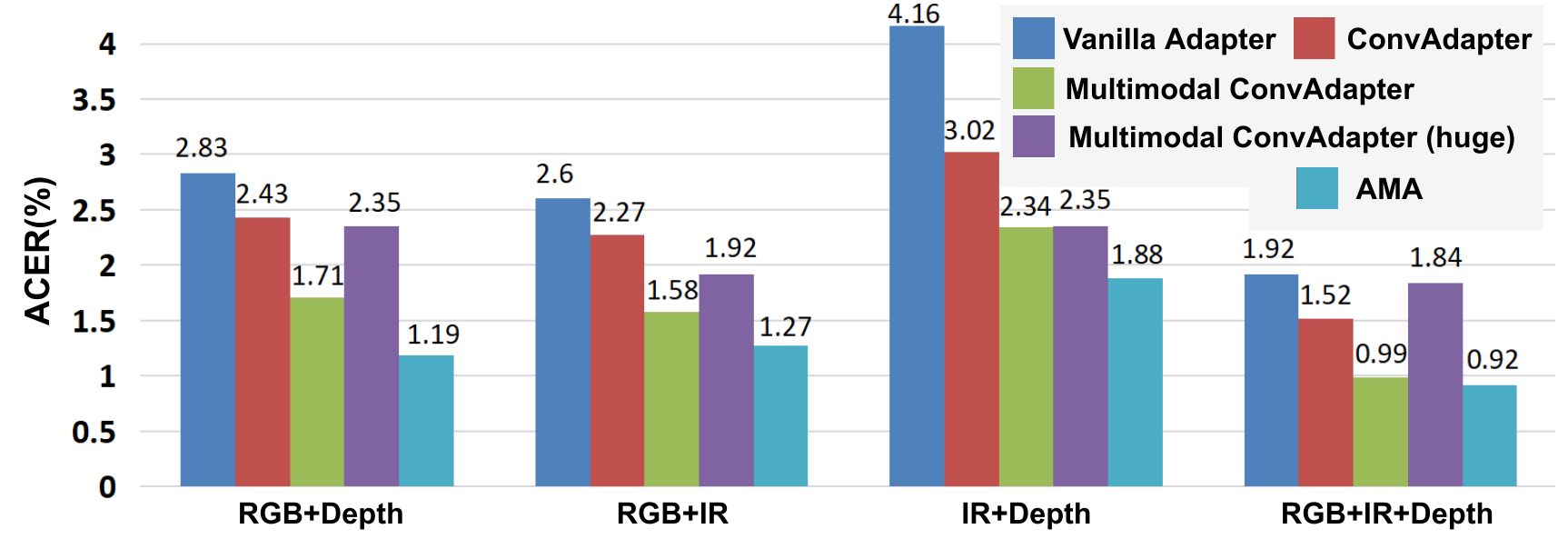}
\vspace{-2.5em}
  \caption{\small{
 Ablation of the adapter types in transformer blocks. }
  }
\label{fig:adapter}
\vspace{-1.3em}
\end{figure}

\subsection{Ablation Study}
\label{sec:ablation}
We also provide the results of ablation studies for inputs with local descriptors and AMA on `seen' protocol of WMCA while studies for M$^{2}$A$^{2}$E on `unseen' protocol of WMCA and cross testing from WMCA to MmFA.

\vspace{0.3em}
\noindent\textbf{Impact of inputs with local descriptors.}\quad   In the default setting of ViT inputs, composition input `GRAY\_HOG\_PLGF' is adopted on for IR modality, while the raw inputs are utilized for RGB and Depth modalities. In this ablation, we consider three local descriptors (`LBP'~\cite{ojala2002multiresolution}, `HOG'~\cite{dalal2005histograms}, `PLGF'~\cite{bhattacharjee2019pattern}) and their compositions (`HOG\_PLGF', `LBP\_HOG\_PLGF', `GRAY\_HOG\_PLGF'). It can be seen from Fig.~\ref{fig:LBP} that the `LBP' input usually performs worse than other features for all three modalities. In contrast, the `PLGF' input achieves reasonable performance (even better performance than raw input for IR modality via direct finetuning). It is clear that raw inputs are good enough for all modalities via ConvAdapter. One highlight is that composition input `GRAY\_HOG\_PLGF' performs the best for IR modality via both direct finetuning and ConvAdapter, indicating the importance of local detailed and illumination invariant cues in IR feature representation.

\vspace{0.3em}
\noindent\textbf{Impact of adapter types.}\quad 
Here we discuss five possible adapter types for efficient multimodal learning, including FC-based `vanilla adapter'~\cite{huang2022adaptive}, independent-modal `ConvAdapter'~\cite{jie2022convolutional}, `multimodal ConvAdapter' with $\Theta_{\text{2D}}$ mapping channels $D'$×$K$ to $D'$, `multimodal ConvAdapter (huge)' with $\Theta_{\text{2D}}$ mapping channels $D'$×$K$ to $D'$×$K$, and adaptive multimodal ConvAdapter (`AMA'). As shown in Fig.~\ref{fig:adapter}, the ConvAdapter based modules perform significantly better than vanilla adapter in multimodal settings, indicating the local inductive biases benefit the ViT-based FAS. Moreover, compared with `ConvAdapter', `multimodal ConvAdapter' reduces more than 0.5\% ACER in all multimodal settings via aggregating multimodal local features. In contrast, we cannot see any performance improvement from `multimodal ConvAdapter (huge)'. In other words, directly learning high-dimensional ($D'$×$K$) convolutional features for all $K$ modalities results in serious overfitting. Compared with `multimodal ConvAdapter', AMA enhances the diversity of features for different modalities via adaptively weighting the shared low-dimensional ($D'$) convolutional features, which decreases 0.52\%, 0.31\%, 1.07\%, 0.07\% ACER for `RGB+Depth', `RGB+IR', `IR+Depth', `RGB+IR+Depth', respectively.

\begin{figure}
\centering
\vspace{-0.8em}
\includegraphics[scale=0.4]{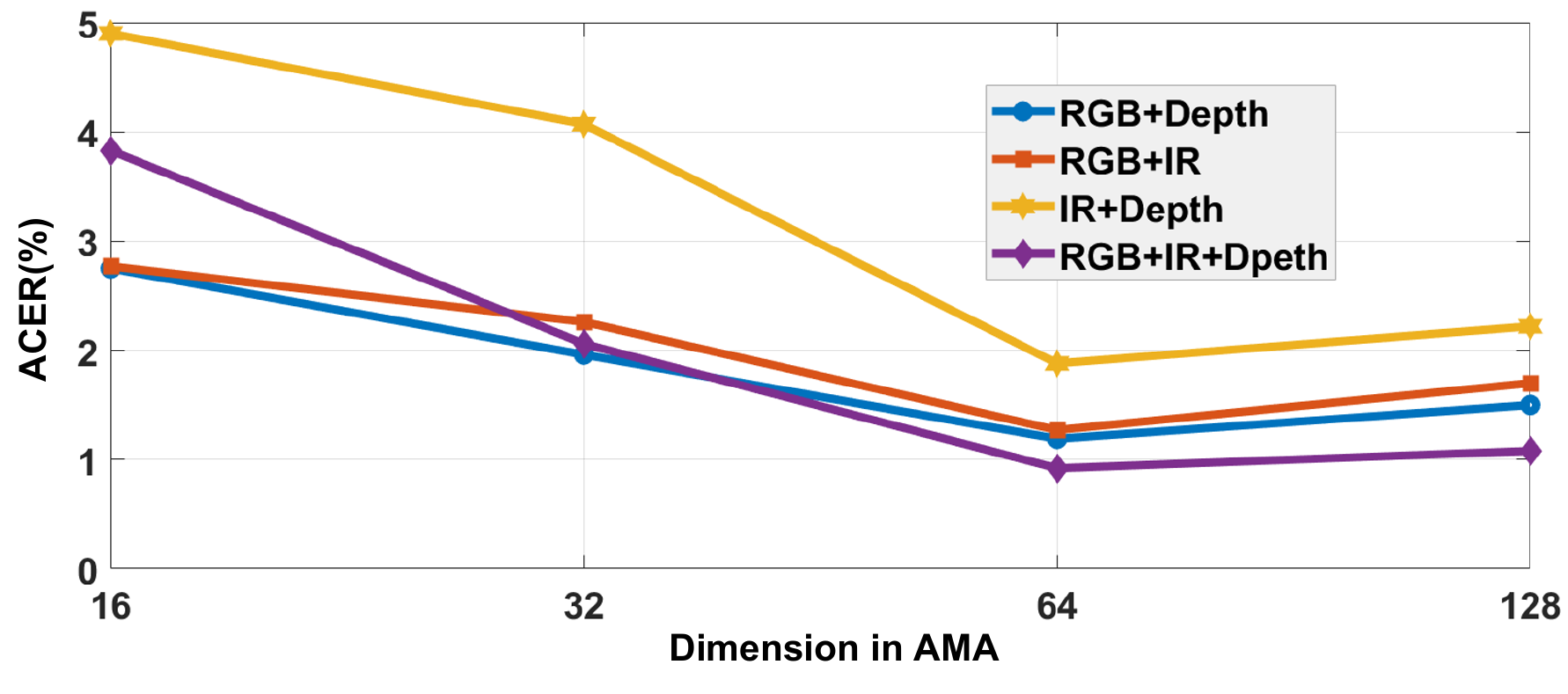}
\vspace{-1.3em}
  \caption{\small{
 Ablation of the hidden dimensions in AMA. }
  }
\label{fig:dimension}
\vspace{-1.3em}
\end{figure}

\begin{figure}
\centering\includegraphics[scale=0.5]{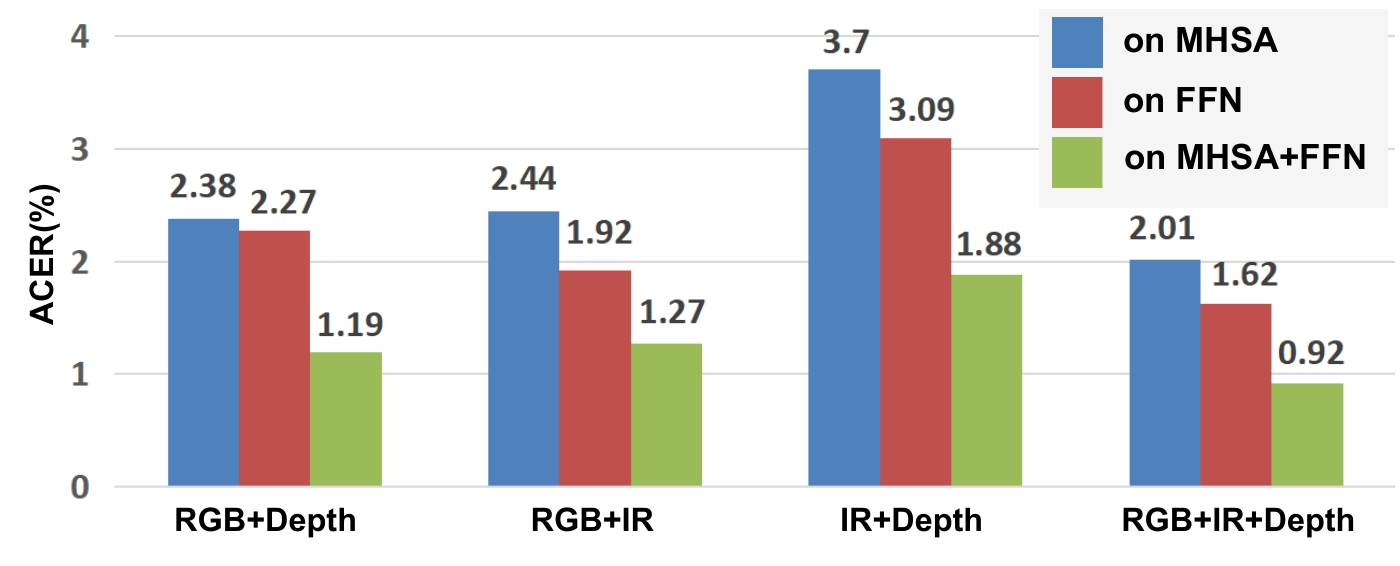}
\vspace{-1.3em}
  \caption{\small{
 Ablation of the AMA positions in transformer blocks. }
  }
\label{fig:position}
\vspace{-1.3em}
\end{figure}

\vspace{0.3em}
\noindent\textbf{Impact of dimension and position of AMA.}\quad 
Here we study the hidden dimensions $D'$ in AMA and the impact of AMA positions in transformer blocks. It can be seen from Fig.~\ref{fig:dimension} that despite more lightweight, lower dimensions (16 and 32) cannot achieve satisfactory performance due to weak representation capacity. The best performance can be achieved when $D'$=64 in all multimodal settings. In terms of AMA positions, it is interesting to find from Fig.~\ref{fig:position} that plugging AMA along FFN performs better than along MHSA in multimodal settings. This might be because the multimodal local features complement the limitation of point-wise receptive field in FFN. Besides, it is reasonable that applying AMA on MHSA+FFN performs the best.

\vspace{0.3em}
\noindent\textbf{Impact of mask ratio in M$^{2}$A$^{2}$E.}\quad
Fig.~\ref{fig:MAE}(a) illustrates the generalization of the M$^{2}$A$^{2}$E pre-trained ViT when finetuning on `unseen' protocol of WMCA and cross testing from MmFA to WMCA. Different from the conclusions from~\cite{he2022masked,bachmann2022multimae} using very large mask ratio (e.g., 75\% and 83\%), we find that mask ratio ranges from 30\% to 50\% are suitable for multimodal FAS, and the best generalization performance on two testing protocols are achieved when mask ratio equals to 40\%. In other words, extreme high mask ratios (e.g., 70\% to 90\%) might force the model to learn too semantic features but ignoring some useful low-/mid-level live/spoof cues.

\vspace{0.3em}
\noindent\textbf{Impact of training epochs and decoder depth in M$^{2}$A$^{2}$E.}\quad 
We also investigate how the training epochs and decoder depth influence the M$^{2}$A$^{2}$E. As shown in Fig.~\ref{fig:MAE}(b) and Fig.~\ref{fig:MAE}(c), training M$^{2}$A$^{2}$E with 400 epochs and decoder of 4 transformer blocks generalizes the best. More training iterations and deeper decoder are not always helpful due to the severe overfits on reconstruction targets.   

\begin{figure}
\centering
\includegraphics[scale=0.52]{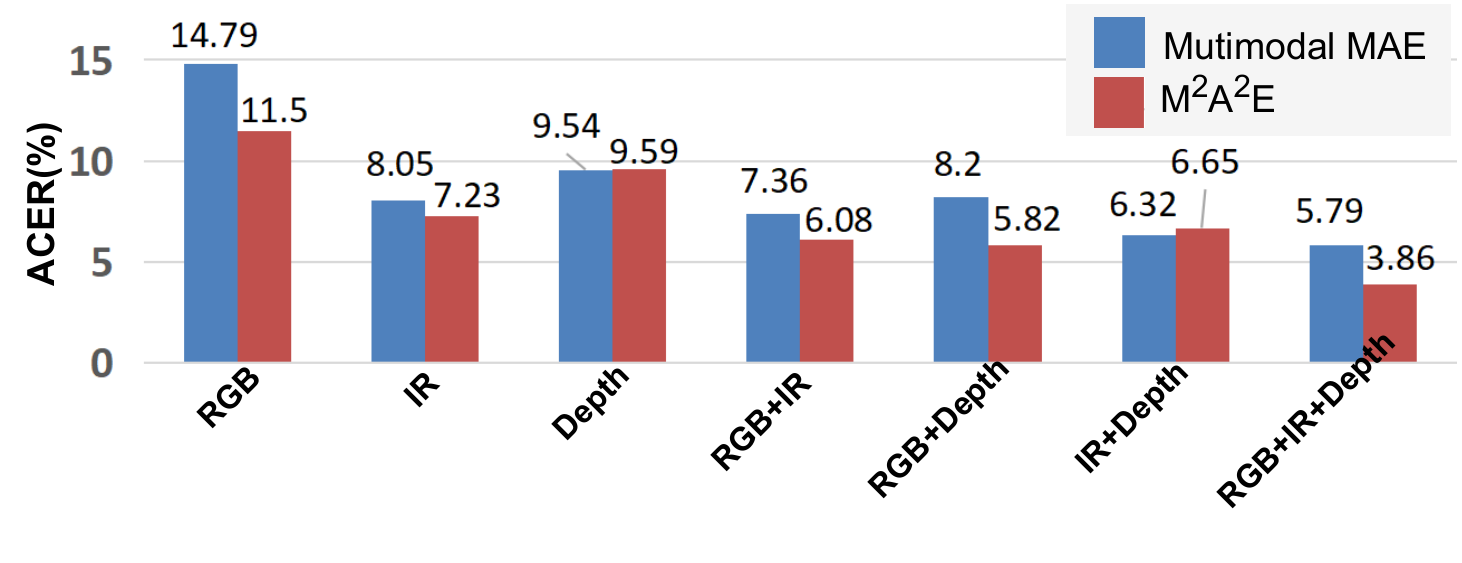}
\vspace{-1.8em}
  \caption{\small{
 Results of the multimodal MAE~\cite{bachmann2022multimae} and our M$^{2}$A$^{2}$E. }
  }
\label{fig:multiMAE}
\vspace{-1.2em}
\end{figure}

\vspace{0.3em}
\noindent\textbf{Comparison between multimodal MAE~\cite{bachmann2022multimae} and M$^{2}$A$^{2}$E.}\quad 
We also compare M$^{2}$A$^{2}$E with the symmetric multimodal MAE~\cite{bachmann2022multimae} when finetuning on all downstream modality settings. It can be seen from Fig.~\ref{fig:multiMAE} that with more challenging reconstruction target (from masked unimodal inputs to multimodal prediction), M$^{2}$A$^{2}$E is outperforms the best settings of multimodal MAE~\cite{bachmann2022multimae} on most modalities (`RGB', `IR', `RGB+IR', `RGB+Depth', `RGB+IR+Depth'), indicating its excellent downstream modality-agnostic capacity.

\begin{figure}
\centering
\vspace{-0.8em}
\includegraphics[scale=0.42]{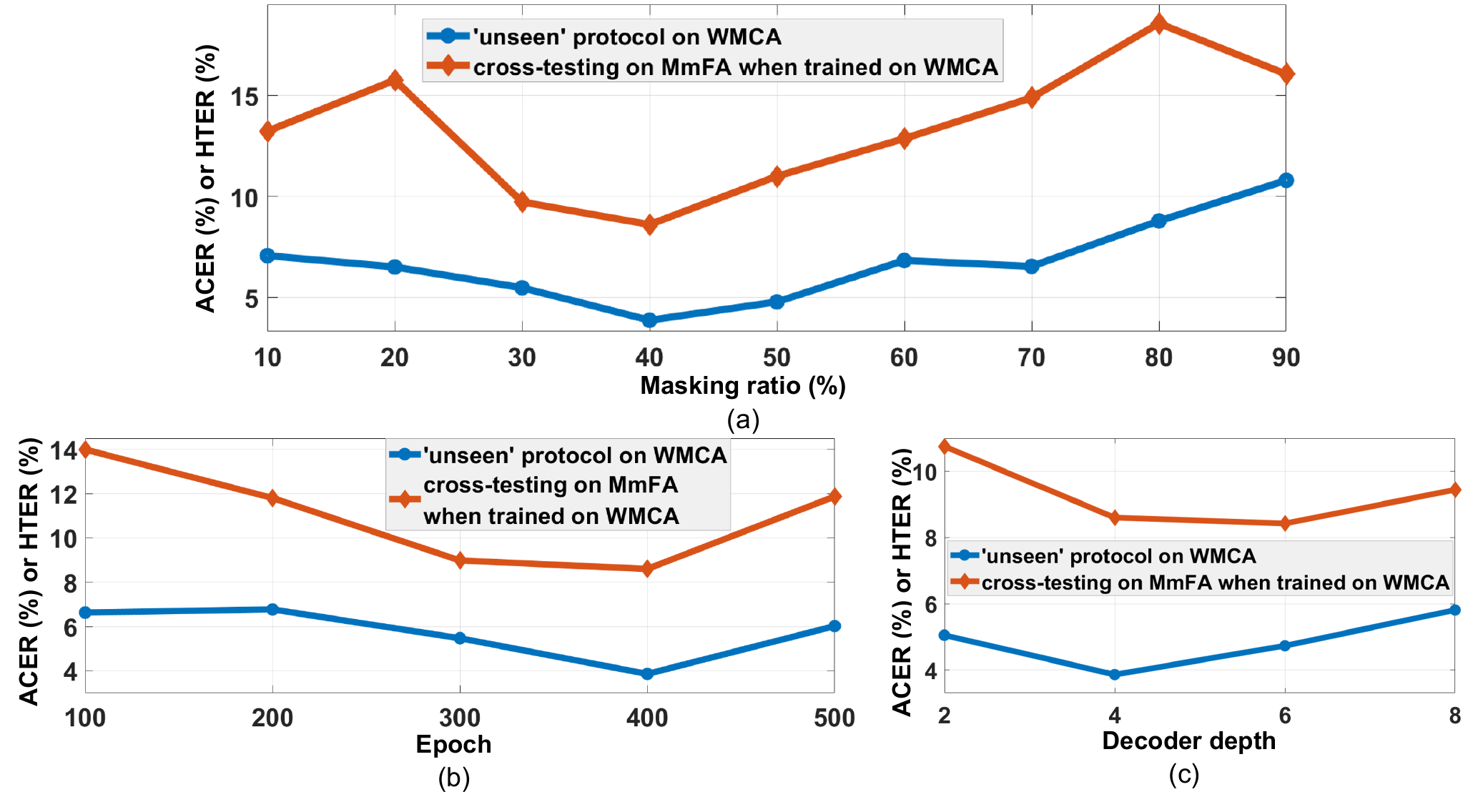}
\vspace{-2.3em}
  \caption{\small{
 Ablation of the (a) mask ratio; (b) self-supervision training epochs; and (c) decoder depth in M$^{2}$A$^{2}$E. }
  }
\label{fig:MAE}
\vspace{-0.8em}
\end{figure}

\vspace{-0.5em}
\section{Conclusions and Future Work}
\vspace{-0.1em}
\label{sec:conc}

In this paper, we investigate three key factors (i.e., inputs, pretraining, and finetuning) for ViT-based multimodal FAS. We propose to combine local feature descriptors for IR inputs, and design the modality-asymmetric masked autoencoder and adaptive multimodal adapter for efficient self-supervised pre-training and supervised finetuning for multimodal FAS. We note that the study of ViT-based multimodal FAS is still at an early stage. Future directions include: 1) Besides inputs, integrating local descriptors into transformer blocks~\cite{yu2022physformer} or adapters is potential for ViT-based multimodal FAS; 2) Besides generalization, the discriminative capacity of M$^{2}$A$^{2}$E pre-trained models should be improved. Some regularization strategies like distillation~\cite{zhang2022mae} might be explored.


\begin{figure*}[t]
\vspace{-0.2em}
\includegraphics[width=1.0\linewidth]{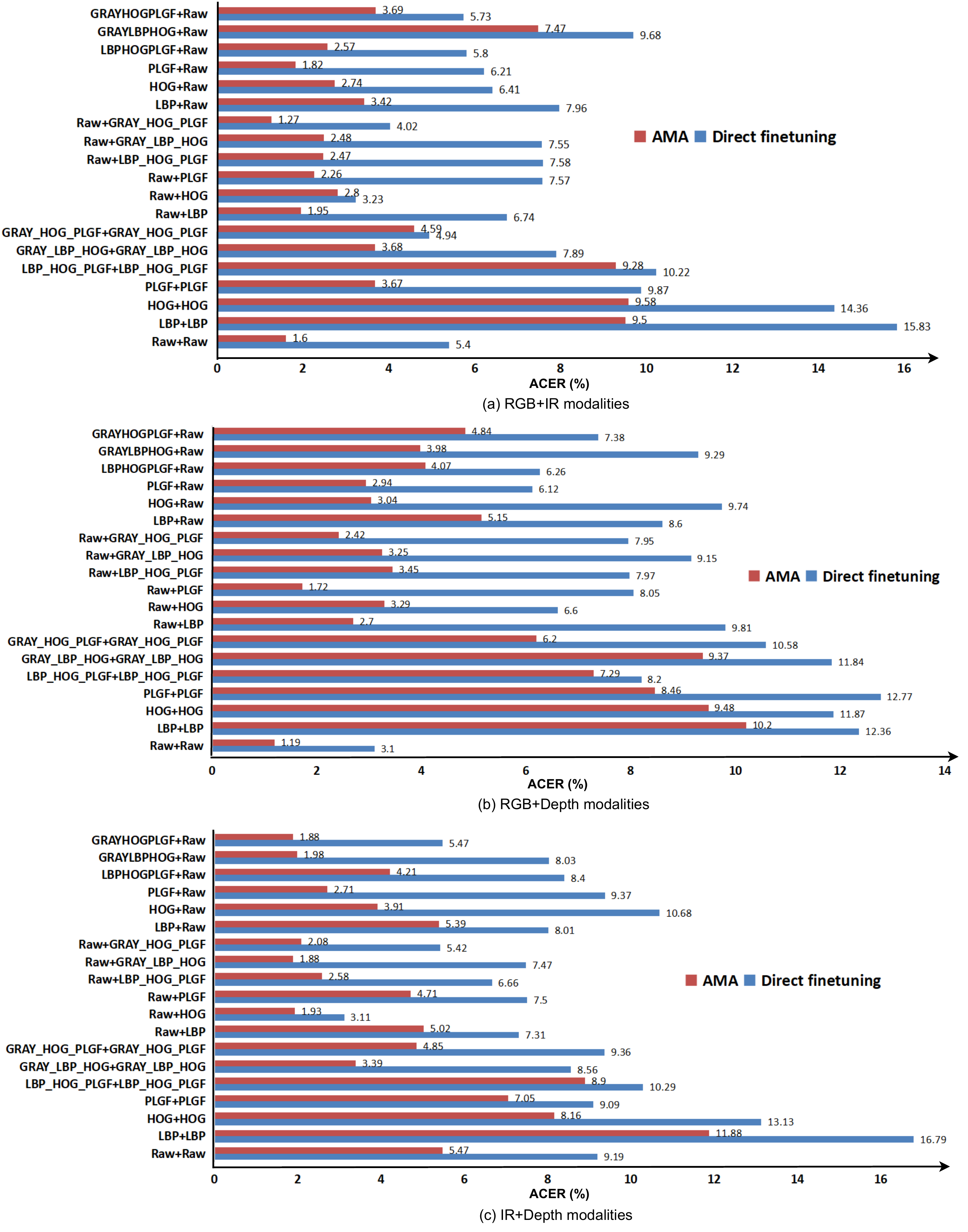}
\centering
\vspace{-1.9em}
\caption{Impacts of bimodal inputs with local feature descriptors (e.g., LBP, HOG, PLGF) for ViT using direct finetuning and adaptive multimodal adapter (AMA) strategies on (a) RGB+IR; (b) RGB+Depth; and (c) IR+Depth modalities. `Raw+HOG' indicates `Raw' input and `HOG' input are used for the first and second modalities, respectively.}
\vspace{-0.3em}
\label{fig:Appendix1}
\end{figure*}

\begin{figure*}[t]
\vspace{-0.2em}
\includegraphics[width=0.95\linewidth]{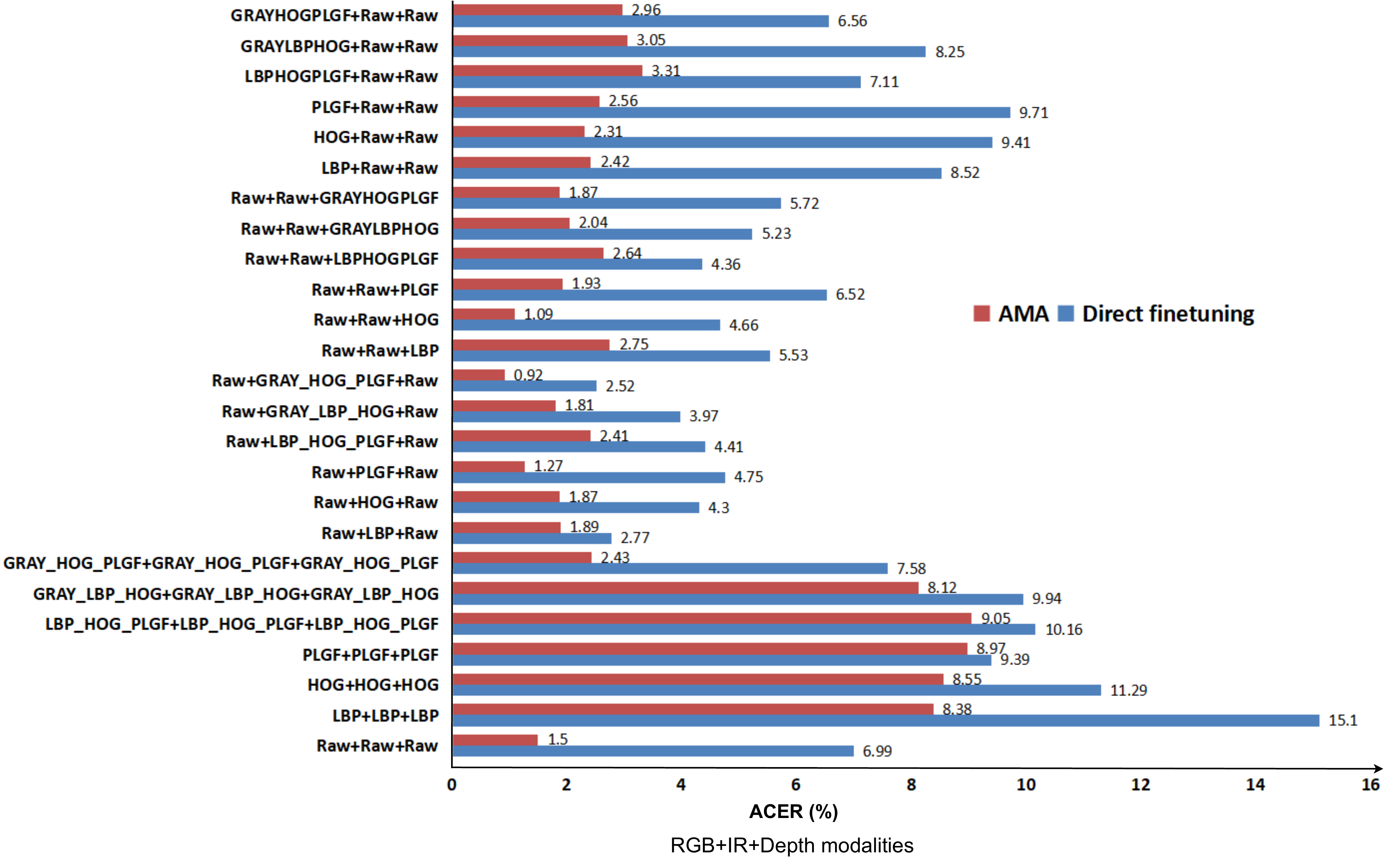}
\centering
\vspace{-1.0em}
\caption{Impacts of trimodal inputs with local feature descriptors for ViT using direct finetuning and AMA strategies on RGB+IR+Depth modalities. `Raw+HOG+Raw' indicates `Raw' input, `HOG' input and `Raw' input are respectively used for three modalities.}
\vspace{-0.3em}
\label{fig:Appendix2}
\end{figure*}

\begin{figure*}[t]
\includegraphics[width=1.0\linewidth]{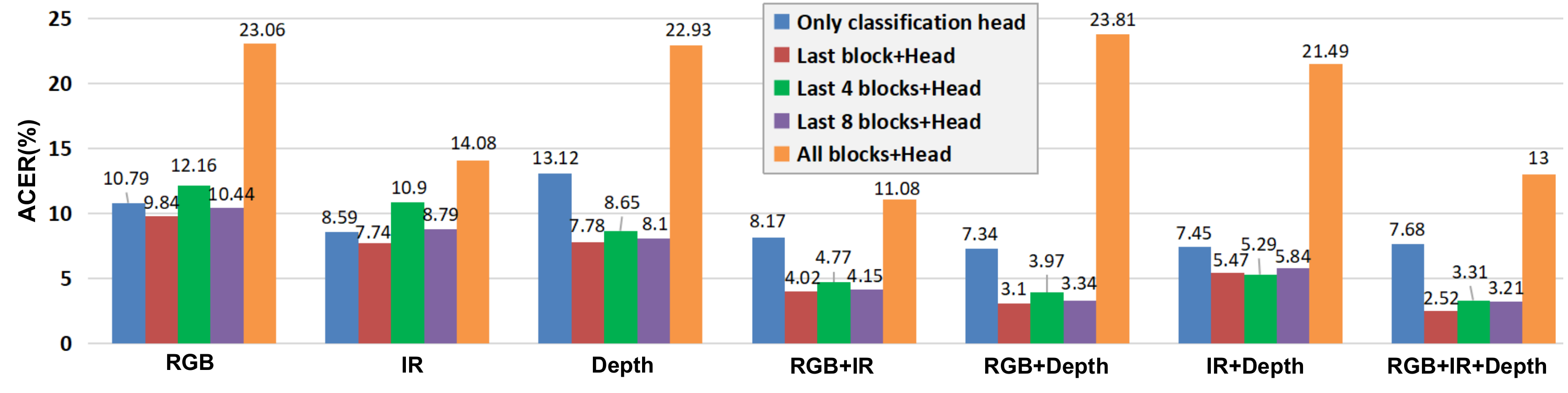}
\centering
\vspace{-1.9em}
\caption{Results of direct finetuning on different transformer blocks.}
\vspace{-0.3em}
\label{fig:Appendix3}
\end{figure*}

\begin{figure*}[t]
\includegraphics[width=1.0\linewidth]{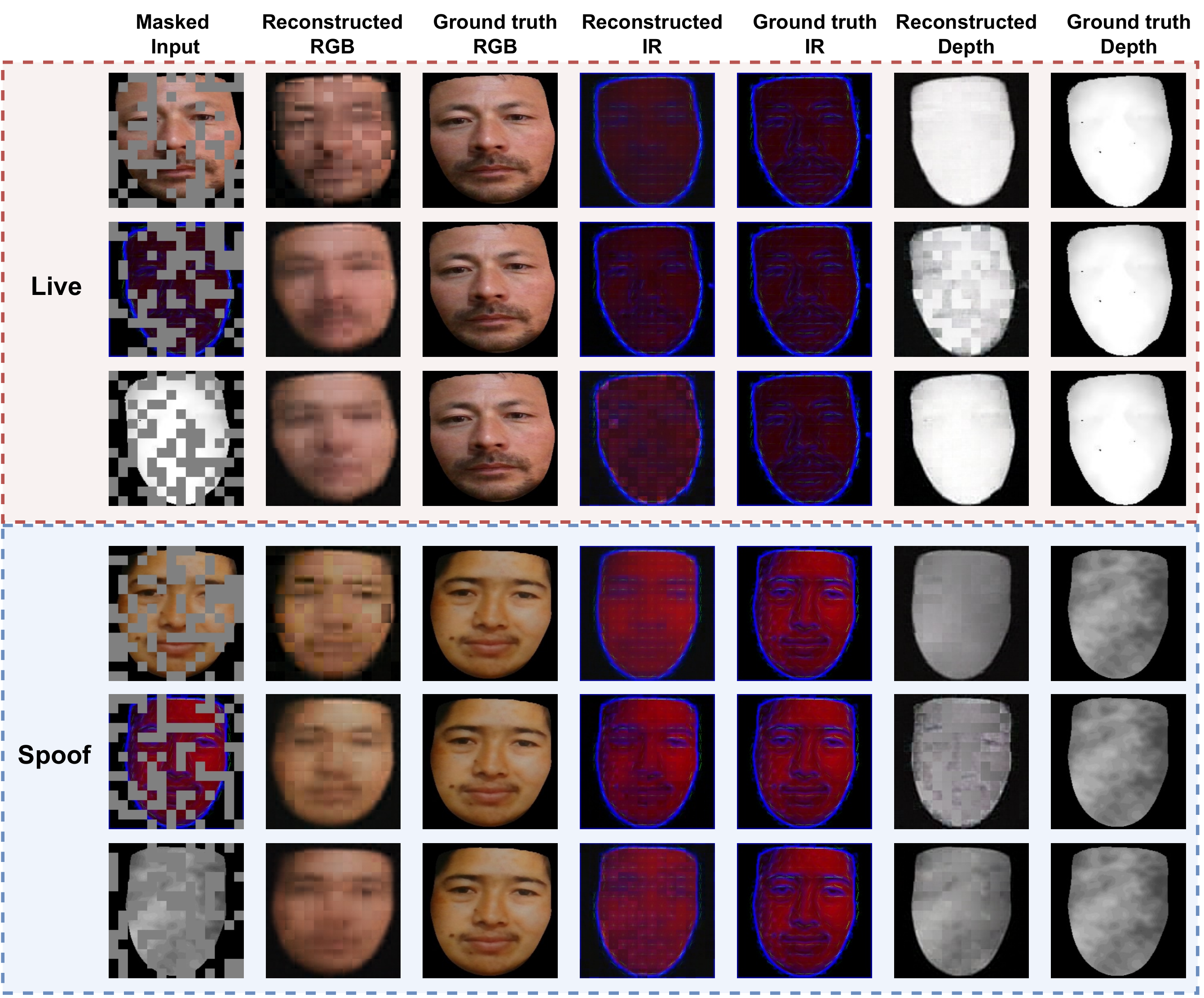}
\centering
\vspace{-1.9em}
\caption{Visualization of the reconstructed multimodal faces from M$^{2}$A$^{2}$E with unimodal masked inputs on CeFA.}
\vspace{-0.3em}
\label{fig:Appendix4}
\end{figure*}

\section*{Appendix A. Ablation on Multimodal Inputs with Local Descriptors }
In the default setting of ViT inputs, composition input `GRAY\_HOG\_PLGF' is adopted on for IR modality, while the raw inputs are utilized for RGB and Depth modalities. Besides the unimodal input results with local descriptors in the main manuscript, we also demonstrate elaborate bimodal and trimodal results with local descriptors on WMCA with Protocol `seen' in Fig.~\ref{fig:Appendix1} and Fig.~\ref{fig:Appendix2}, respectively. Some observations can be found: 1) in the multimodal settings (RGB+IR, IR+Depth, and RGB+IR+Depth), the composition input `GRAY\_HOG\_PLGF' for IR modality with raw inputs for other modalities performs the best via both direct finetuning and AMA; and 2) leveraging the local descriptor inputs for IR modality via direct finetuing, the performance on `IR+Depth' and `RGB+IR+Depth' are significantly improved compared with using all raw inputs. They all indicate the importance of local detailed and illumination invariant cues in IR feature representation.

\section*{Appendix B. Ablation on Direct Finetuning}
Here we discuss five finetuning strategies for ImageNet pretrained ViT-Base including finetuning 1) all transformer blocks and classification head; 2) last 8 transformer blocks and classification head; 3) last 4 transformer blocks and classification head; 4) last transformer block and classification head; and 5) only classification head. The results on WMCA with Protocol `seen' are shown in Fig.~\ref{fig:Appendix3}. The composition input `GRAY\_HOG\_PLGF' is used for IR modality. It is clear that finetuning all transformer blocks perform the worst due to the overfitting issues in huge trainable parameters. In contrast, finetuning the last transformer block or last 8 transformer blocks can achieve stable and reasonable performance. We adopt the strategy of finetuning last transformer block and classification head as the defaulted direct finetuning setting due to its efficiency.

\section*{Appendix C. Visualization of M$^{2}$A$^{2}$E}
We also conduct experiments to visual the reconstruction results of the proposed M$^{2}$A$^{2}$E. As illustrated in Fig.~\ref{fig:Appendix4}, we find that even though with unimodal (40\%) masked input, the reconstruction of all three modalities still keep some important live/spoof clues (e.g., convincing facial geometric depth predicted from masked RGB input for live samples while equivocal depth info for spoof ones), which benefits the self-supervised pre-training for generalized task-aware feature representation. 

{\small
\bibliographystyle{ieee_fullname}
\bibliography{egbib}
}

\end{document}